\documentclass[letterpaper, 10 pt, conference]{ieeeconf}  

\IEEEoverridecommandlockouts                              
                                                 
\usepackage{graphics} 
\usepackage{epsfig} 
\usepackage{mathptmx} 
\usepackage{times} 
\usepackage{amsmath} 
\usepackage{amssymb}  
\usepackage{url}
\usepackage{romannum}
\usepackage{xcolor}
\usepackage{float}
\usepackage{url}
\usepackage[ruled,vlined]{algorithm2e}
\usepackage{array}
\usepackage{placeins}
\usepackage{adjustbox}
\usepackage{bigstrut}
\usepackage{adjustbox}

\title{\LARGE \bf

Data-Efficient Sequence-Based Visual Place Recognition with Highly Compressed JPEG Images

}

\author{Mihnea-Alexandru Tomiță$^{1}$, Bruno Ferrarini$^{1}$, Michael Milford$^{2}$, Klaus McDonald-Maier$^{1}$, Shoaib Ehsan$^{1,3}$
\thanks{$^{1}$Authors are with the School of Computer Science and Electronic Engineering, University of Essex, CO4 3SQ, United Kingdom
        {\tt\small matomi@essex.ac.uk, bferra@essex.ac.uk, kdm@essex.ac.uk, sehsan@essex.ac.uk}. }
\thanks{$^{2}$Michael Milford is with the School of Electrical Engineering and Computer Science, Queensland University of Technology, Brisbane, QLD 4000, Australia
        {\tt\small michael.milford@qut.edu.au}}
\thanks{$^{3}$Shoaib Ehsan is also with the School of Electronics and Computer Science, University of Southampton, SO17 1BJ, United Kingdom
        {\tt\small s.ehsan@soton.ac.uk}}
\thanks{This work is supported by the UK Engineering and Physical Sciences Research Council through grants EP/R02572X/1 and EP/P017487/1.}
}

\begin{document}

     \maketitle
    \thispagestyle{empty}
    \pagestyle{empty}

     \begin{abstract}

     Visual Place Recognition (VPR) is a fundamental task that allows a robotic platform to successfully localise itself in the environment. For decentralised VPR applications where the visual data has to be transmitted between several agents, the communication channel may restrict the localisation process when limited bandwidth is available. JPEG is an image compression standard that can employ high compression ratios to facilitate lower data transmission for VPR applications. However, when applying high levels of JPEG compression, both the image clarity and size are drastically reduced. In this paper, we incorporate sequence-based filtering in a number of well-established, learnt and non-learnt VPR techniques to overcome the performance loss resulted from introducing high levels of JPEG compression. The sequence length that enables 100\% place matching performance is reported and an analysis of the amount of data required for each VPR technique to perform the transfer on the entire spectrum of JPEG compression is provided. Moreover, the time required by each VPR technique to perform place matching is investigated, on both uniformly and non-uniformly JPEG compressed data. The results show that it is beneficial to use a highly compressed JPEG dataset with an increased sequence length, as similar levels of VPR performance are reported at a significantly reduced bandwidth. The results presented in this paper also emphasize that there is a trade-off between the amount of data transferred and the total time required to perform VPR. Our experiments also suggest that is often favourable to compress the query images to the same quality of the map, as more efficient place matching can be performed. The experiments are conducted on several VPR datasets, under mild to extreme JPEG compression.
    

      \end{abstract}

      
      \section{Introduction}\label{introduction}
      
      In recent years, considerable effort has been devoted to creating decentralised VPR pipelines \cite{cieslewski2017efficient,cieslewski2018data,burguera2020unsupervised} that greatly benefit applications such as search and rescue, where several robotic platforms are required to successfully map, localise and navigate through the environment. Visual Place Recognition (VPR), part of the Simultaneous Localisation and Mapping (SLAM), is an essential task for the localisation process, where each robotic platform is required to successfully navigate through the environment using the visual information gathered from the on-board camera. For these applications to be successful, the visual data has to be transmitted between each agent, in order to facilitate swift collaboration between them. Moreover, when working with a narrow bandwidth, it is not always feasible to transfer a large amount of data. 
      
       \begin{figure}[t]
            \centering
            \begin{tabular}{ c }

                \includegraphics[width=230pt, trim=2 8 8 8, clip]{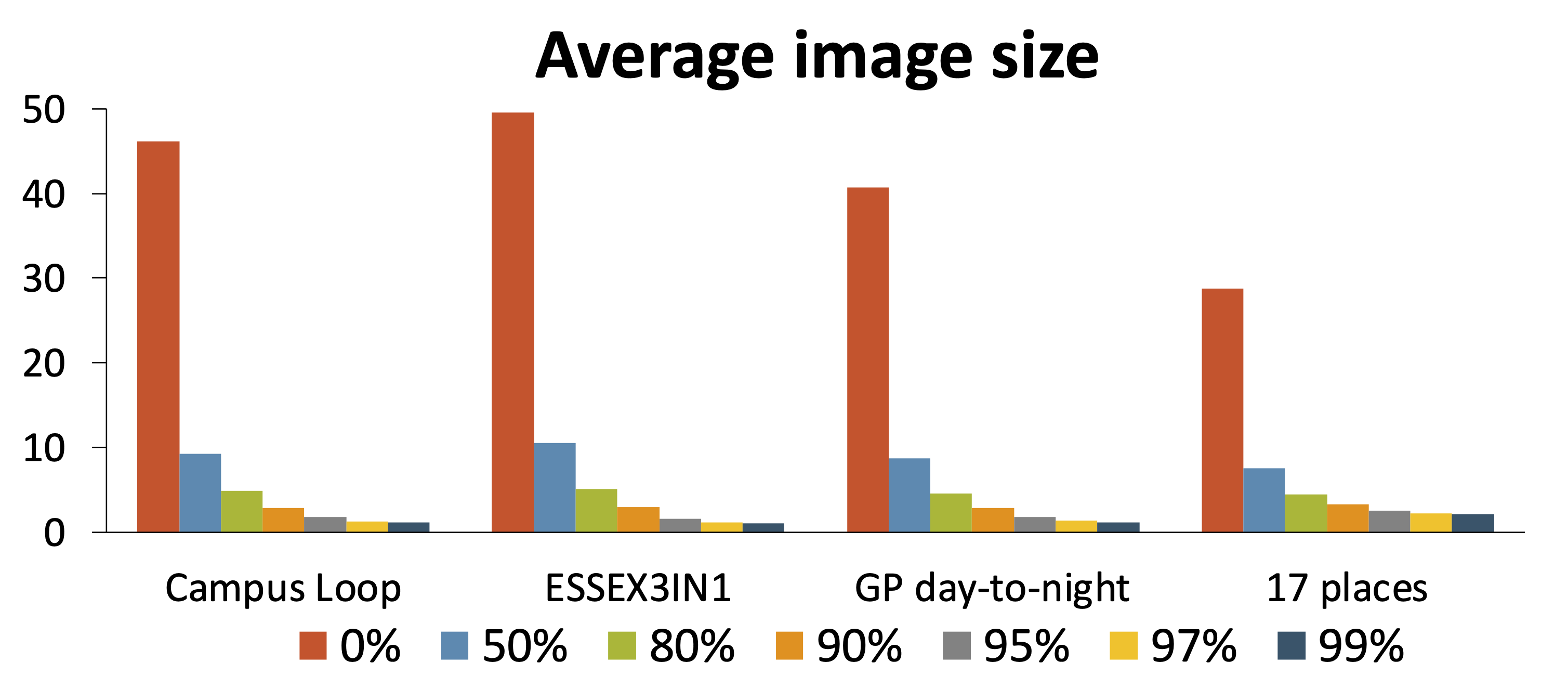} 
                
            \end{tabular}
            \caption{The average image size in Kilobytes (KB) taken from each dataset with multiple JPEG compression ratios applied.}
            \label{averageimagesize}
            \end{figure}
      
      
      
      JPEG \cite{hudson_jpeg-1_2018} is a lossy image representation technique capable of drastically reducing both the storage and transmission requirements of visual data. As JPEG can employ high compression ratios when compared to other techniques \cite{8301939,mateika2007analysis}, we believe that it can effectively lower the amount of data transferred in distributed VPR scenarios, as image size can be prioritised over image quality. 
      Because JPEG is designed to have a minimal impact on the human perception system \cite{haines1992effects}, the VPR performance is drastically reduced especially when working with highly JPEG compressed images. Recently in \cite{9849680} it has been shown that by introducing sequence-based filtering on top of single-frame based VPR techniques, their place matching performance is greatly improved. In this paper, to compensate for the performance degradation resulted from utilising high ratios of JPEG compression, we propose to introduce sequence-based filtering in several well-established VPR techniques. Moreover, the total time required to perform VPR is analysed to determine whether it is more efficient to compress the query images at the same quality as the map. In summary, our contributions are the following:
      
      
      \begin{itemize}
      \item The application of sequence-based filtering is investigated on highly JPEG compressed data. 
      An analysis of the sequence length that enables perfect place matching performance and the amount of data required to transmit for each VPR technique is provided. This study is performed on several datasets containing illumination, viewpoint and seasonal variations, accurately depicting the most widely encountered changes in the environment. 
      \item The time required for each VPR technique to perform place matching is investigated throughout the entire spectrum of JPEG compression. An analysis is performed on the uniformly and non-uniformly JPEG compressed ESSEX3IN1 dataset that suggests that both the query and map images should be compressed at the same ratio, as it facilitates more efficient VPR at reduced bandwidth.
      \end{itemize}
      
      The remainder of this paper is organised as follows: Section \ref{literature_review} presents the literature review. In section \ref{methodology}, JPEG compression is presented together with the implementation of sequence-based filtering. Section \ref{experimental_study} describes the selection of VPR techniques, datasets utilised and the performance metric employed for performing the analysis. Section \ref{results} presents the detailed results and analysis. Finally, the conclusions are presented in Section \ref{conclusion}.

      \section{Literature Review}   \label{literature_review}
     
      
      Before CNNs emerged in popularity, handcrafted feature descriptors such as Scale-Invariant Feature Transform (SIFT) \cite{SIFT,lowe2004distinctive} and Speeded-Up Robust Features (SURF) \cite{SURF} were primarily utilised to facilitate VPR tasks \cite{se2002mobile}, \cite{andreasson2004topological}, \cite{stumm2013probabilistic}, \cite{kovsecka2005global}, \cite{murillo2007surf}. FAB-MAP (Frequent Appearance Based Mapping) \cite{cummins2011appearance} is an appearance based place recognition system that can successfully perform loop-closure detection and deal with perceptual aliased images. This VPR technique represents places as words and utilises SURF to detect the features that are present in images. CAT-SLAM \cite{maddern2012cat} extends FAB-MAP by introducing odometry calculations. FrameSLAM \cite{konolige2008frameslam} employed Center Surround Extremas (CenSurE) \cite{agrawal2008censure} as it facilitates real-time detection and matching of image features. Vector of Locally Aggregated Descriptors (VLAD) \cite{jegou2010aggregating} and the Bag-of-Words model (BoW) \cite{4270197} build an image descriptor of fixed length by aggregating local feature descriptors around centroids. The work presented by the authors of \cite{8792942} shows how VLAD has been applied to VPR. Gist \cite{globalimagefeatures}, \cite{oliva2001modeling} is a global feature descriptor that has been employed by the authors of \cite{murillo2009experiments, singh2010visual,siagian2009biologically} to perform VPR. Histogram-of-Oriented-Gradients (HOG) \cite{dalal2005histograms} is a computationally efficient whole-image descriptor, designed to handle appearance changes in the environment \cite{zaffar2021vpr}. However, there is no universal VPR descriptor that can handle every environmental change as presented in \cite{8968579,zaffar2019levelling}. To overcome these limitations, the authors of \cite{SwitchHit} proposed a switching system based on complementary of several VPR techniques \cite{9459537}. The resulting system is able to select the best possible descriptor that can handle a particular place or environmental condition.
     
     Chen \textit{et al.} studied the suitability of deep-learning for VPR tasks in \cite{chen2014convolutional}. The authors combined all the 21 layers of the Overfeat network \cite{sermanet2013overfeat} trained on ImageNet 2012 dataset with both the spatial and sequential filter of SeqSLAM \cite{milford2012seqslam}. AmosNet and HybridNet \cite{chen2017deep} are two neural network architectures trained on the Specific PlacEs Dataset (SPED). NetVLAD \cite{arandjelovic2016netvlad} introduces a new layer based on a generalised VLAD descriptor, which is highly robust to viewpoint variations. A light-weight CNN-based technique robust to environmental changes is presented in \cite{khaliq2019camal}. Moreover, the proposed technique has low resource utilisation. CALC \cite{merrill2018lightweight} is a CNN-based VPR descriptor robust to viewpoint and illumination variations. Cross-Region-Bow \cite{chen2017only} is a VPR technique tolerant to viewpoint variations. It searches for regions of interest (ROIs) and uses the convolutional layer to create the representation of these salient regions. RegionVLAD \cite{khaliq2019holistic} is a light-weight CNN-based VPR technique, capable of detecting salient features and filtering confusing elements. Both RegionVLAD and Cross-Region-Bow are based on a similar approach, however RegionVLAD uses VLAD for feature pooling. AlexNet ConvNet \cite{krizhevsky2012imagenet} pre-trained on the ImageNet ILSVRC dataset \cite{russakovsky2015imagenet} has been utilised for object recognition in \cite{AlexNet}. Binary neural networks (BNNs) have recently been proposed in \cite{9725251} and \cite{HDBS} for VPR applications as they require less computational power to achieve comparable levels of performance as full-precision systems. Notwithstanding, BNNs require an inference engine or dedicated hardware which allows them to perform efficient computations of bitwise operations. In \cite{9672749}, the authors propose an efficient and lightweight neural network based on the drosophila neural system to address VPR.
     
     In \cite{9849680}, a study is performed where sequence-based filtering is introduced in several single-frame-based VPR descriptors, highlighting the benefits and drawbacks. SeqSLAM \cite{milford2012seqslam} performs VPR utilising a sequence of constant length to determine a robot's position in the environment. SMART \cite{pepperell2014all} extended SeqSLAM by incorporating the odometry into its calculations. In \cite{naseer2014robust}, robust localisation is performed utilising a sequence-based VPR system that is capable of handling substantial seasonal change. The authors of ConvSequential-SLAM \cite{tomita2021convsequential} proposed a sequence-based VPR system utilising an adaptive sequence-based matching approach to address VPR in dynamic environments. In \cite{chancan2020deepseqslam}, the authors propose a trainable CNN+RNN architecture for sequence-based place recognition, entitled DeepSeqSLAM. 
    
     Decentralized VPR is a method of identifying previously visited places utilising the visual information gathered by a distributed system, rather than relying on a centralised server that performs the place matching computations \cite{cieslewski2017efficient, 7811208}. This is especially important in applications such as search and rescue, where each agent can cover a part of the environment, resulting in a swifter task completion. Moreover, in similar scenarios a centralized point of failure or dependence may be undesirable. The authors of \cite{cieslewski2018data} presented a data-efficient decentralised visual SLAM system where the data association scales linearly with the number of agents that are part of the multi-robot SLAM system. In \cite{burguera2020unsupervised}, the authors introduce a loop detection system for multi-robot underwater visual SLAM applications. The authors of \cite{7811208} propose a decentralised system which scales in a similar fashion to a centralised approach by distributing the query images into multiple clusters. For multi-robot SLAM applications where the visual data has to be shared between several agents, it is often the case that the transmission process is proven to be difficult due to limited bandwidth, as identified in \cite{cieslewski2017efficient, 7811208, cieslewski2018data, burguera2020unsupervised}. For this reason, this paper analyses the data transfer problem in VPR utilising highly JPEG compressed images as they are smaller in size when compared to an image descriptor (refer to section \ref{results}).
    
     
    
     
      \section{Methodology} \label{methodology}
      
      \subsection{JPEG Compression}
      
      JPEG is a lossy compression method utilised for digital images that allows users to select and adjust the amount of compression applied to any given image. Depending on the selected JPEG compression amount, the visual quality of the image is not compromised when mild image compression is selected (e.g. 50\% JPEG compression). The compression process is accomplished by removing small variations in color and texture, which ultimately would have a minimal impact on the human perception system \cite{haines1992effects}. On the contrary, when extremely compressing an image (e.g. 99\% JPEG compression) the amount of data required to represent the image is drastically reduced, thus also altering its visual aspect \cite{hudson_jpeg-1_2018}. The compression parameter of a JPEG compression function has values in range [0,99] and is responsible for adjusting the amount of information that is removed during the compression process. A lower compression ratio would result in a higher image quality and an increased file size. Conversely, when a high value is assigned to the compression parameter, the clarity of the image is drastically reduced and it only occupies a fraction of its original size (as observed in Fig. \ref{averageimagesize}). For this reason, JPEG compression can benefit decentralised VPR applications as less visual data is required to be transmitted. It is important to note that JPEG compression only affects the image's quality and file size. 
      

      \subsection{Implementation of Sequence-Based Filtering}
        In the single-image-based approach, each query image is matched against every reference image in the map. The cosine \cite{chen2017only} is utilised to generate a similarity score between each query-reference pair, whose values are in range [0,1]. Thus, the reference image that has the highest score is retrieved as the best representation for any given query image.  
        
        In contrast with the previously mentioned matching schema, sequence-based filtering searches entire sequences of query and reference images \cite{9849680}. A sequence of consecutive images of length \textbf{K} is created for every query image \textit{$q_i$} as follows:
        \begin{equation}
        q_{seq} =
            \begin{bmatrix}
                q_{i} & q_{i+1} & q_{i+2} & \dots  & q_{K}
            \end{bmatrix}
            \label{eq:querysequence}
       \end{equation}
        
        For each $q_{seq}$ created, a similar list of consecutive reference images $r_{seq}$ is generated for every reference image \textit{$r_i$} in the map. 
        
        During the matching process, each query image \textit{$q_i$} from $q_{seq}$ is matched with its corresponding reference image \textit{$r_i$} from $r_{seq}$, utilising the cosine to generate a similarity score \textit{$s_i$}. Furthermore, to get the matching score for an entire sequence of images of length \textbf{K}, the arithmetic mean of every \textit{$s_i$} in the sequence analysed is calculated as follows: 
        
        \begin{equation}
        score = \frac{s_{i} + s_{i+1} + s_{i+2} + ...  + s_{K}}{K}
            \label{eq:score}
       \end{equation}

        Further details on the implementation of both the single-image-based and sequence-based filtering have been discussed at length in \cite{9849680} and are not provided here to avoid redundancy.

      \section{Experimental Setup} \label{experimental_study}
      
      \subsection{VPR Techniques}\label{vprtechniques}
     
     A selection of four well-established, learnt and non-learnt VPR techniques have been employed in this work including: NetVLAD \cite{arandjelovic2016netvlad}, HybridNet \cite{chen2017deep}, RegionVLAD \cite{khaliq2019holistic} and HOG \cite{dalal2005histograms}. These techniques have been selected as they span over a broad spectrum, ranging from descriptors that are designed for seasonal and illumination variations such as HybridNet, and descriptors that achieve viewpoint tolerances such as NetVLAD. Due to the agnostic nature of the sequence-based filtering schema proposed in \cite{tomita2021convsequential}, this approach can be introduced in every VPR technique mentioned above. For this reason, this particular matching schema is utilised as it enables a fair comparison of each sequence-based VPR technique on highly JPEG compressed data.
     
     
      
      \begin{figure*}[t]
            \centering
            \begin{tabular}{ c c }

                \includegraphics[width=245pt, trim=8 8 8 8, clip]{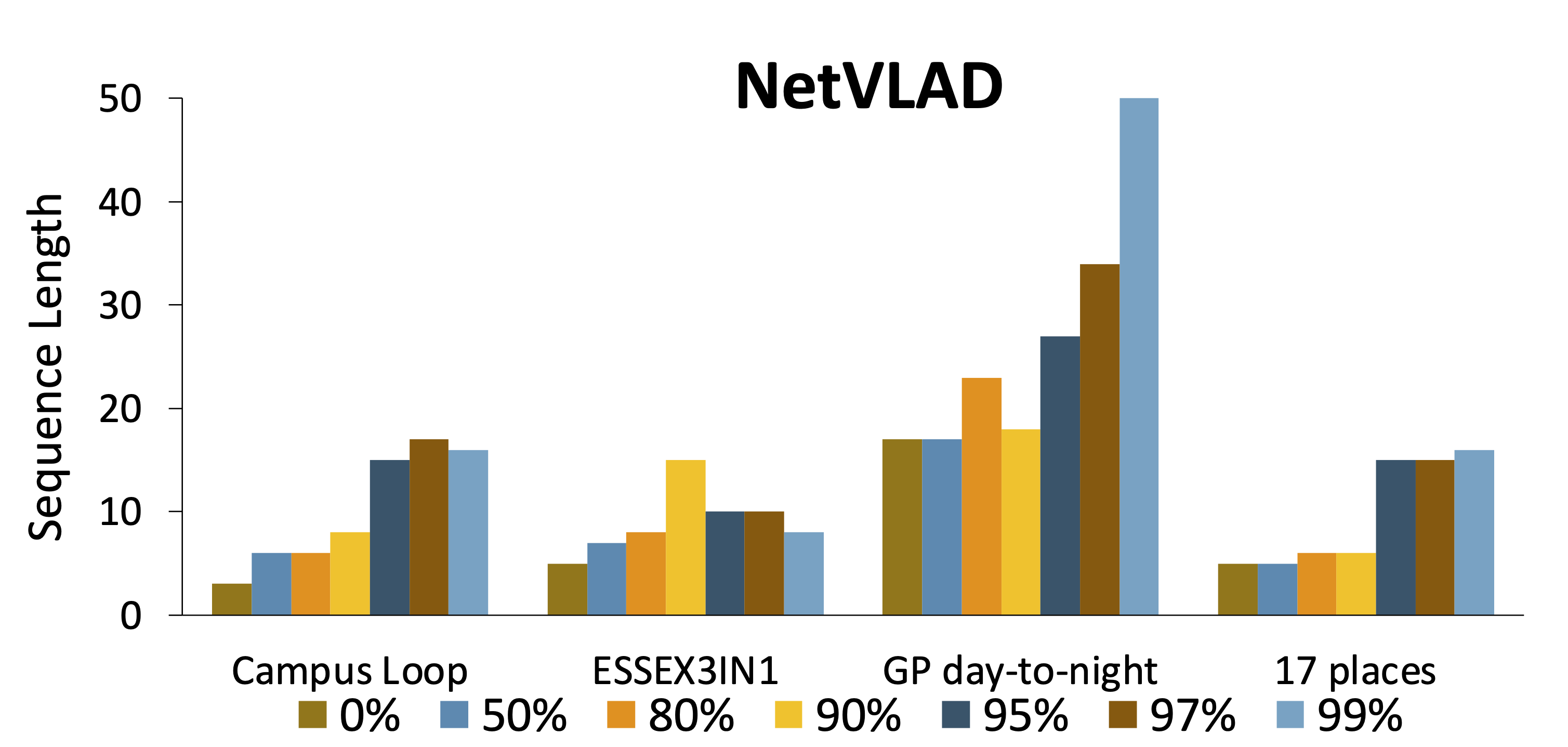} & 
                \includegraphics[width=245pt, trim=8 8 8 8, clip]{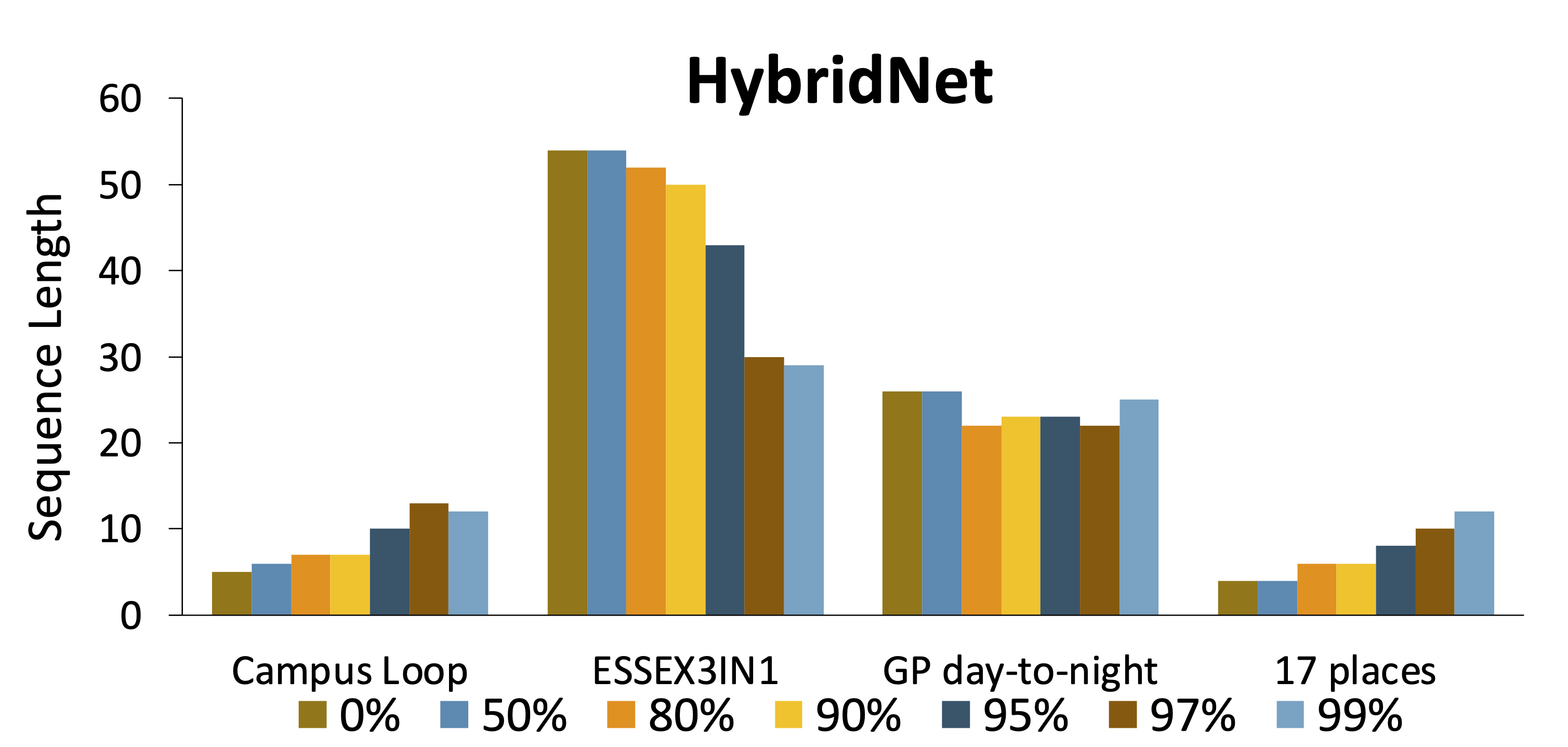} \\
                \includegraphics[width=245pt, trim=8 8 8 8, clip]{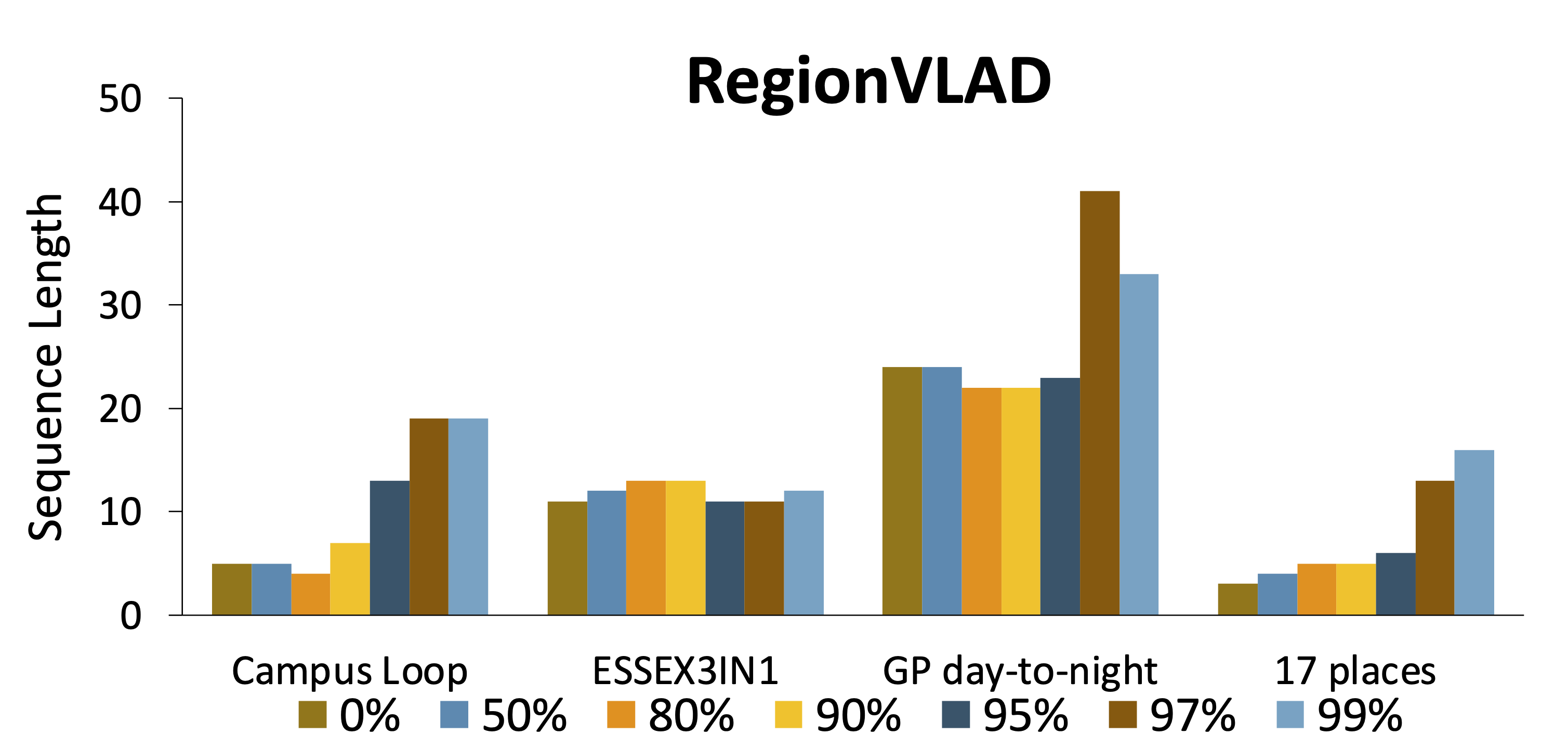} &
                \includegraphics[width=245pt, trim=8 8 8 8, clip]{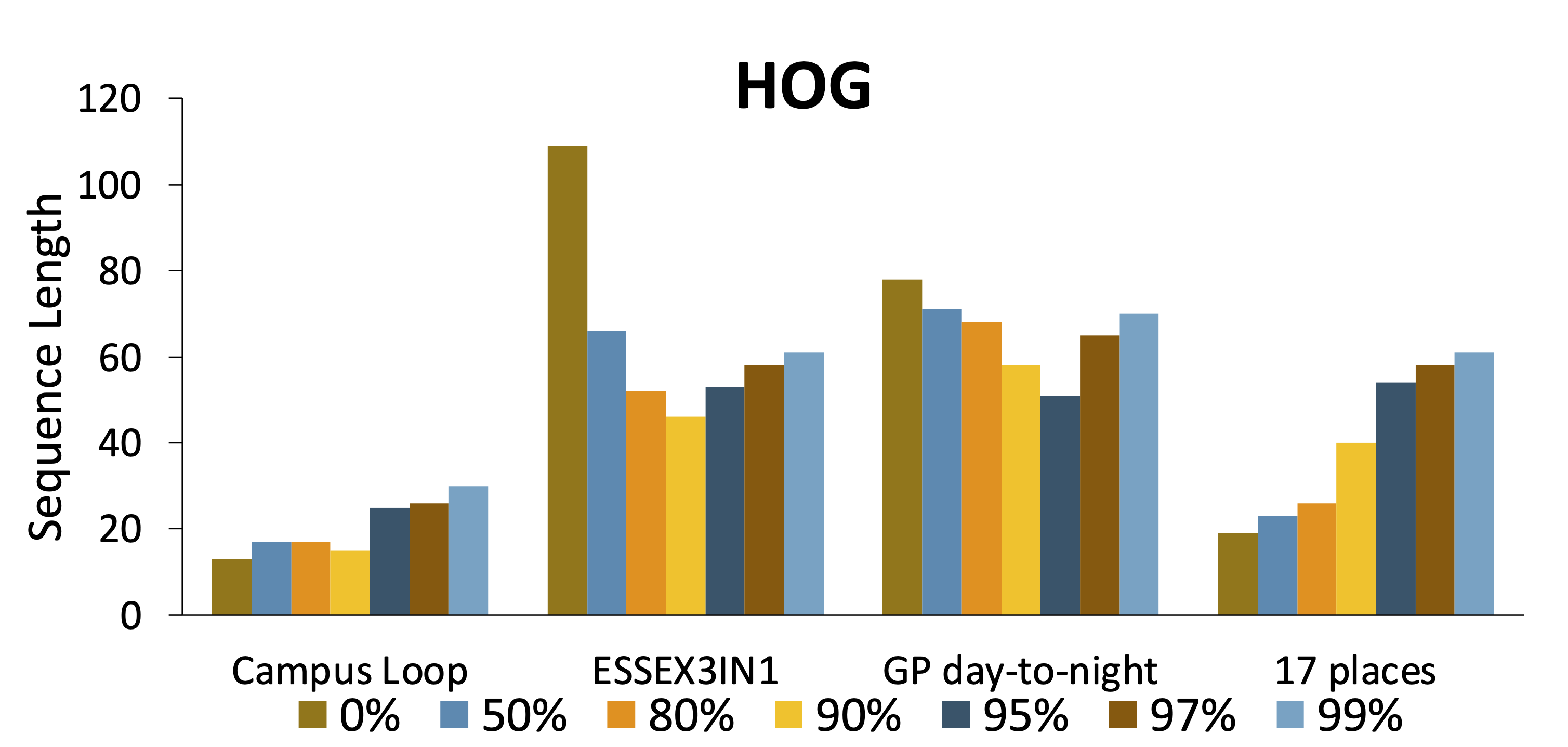}
                
            \end{tabular}
            \caption{The sequence length required to reach maximum accuracy for every JPEG compression ratio.}
            \label{sequencelengthgraph}
            \end{figure*} 
      
     \subsection{Test Datasets}
     
     This paper employs a total of four datasets widely used by the VPR community that present different scenarios where the environment is affected by illumination, viewpoint and/or seasonal variations. The first dataset utilised in our work is Campus Loop \cite{merrill2018lightweight}, that contains a total of 200 images, equally split in 100 query and 100 reference images. This is a challenging dataset due to presence of viewpoint and seasonal variations. ESSEX3IN1 \cite{ESSEX3IN1} has 210 query and 210 reference images that include perceptual aliasing and confusing places. The third dataset employed in this work is Gardens Point (GP) day-to-night \cite{sunderhauf2015performance}, that contains 200 query images (\textit{day left}) and 200 reference images (\textit{night right}). This dataset poses several challenges to VPR applications such as viewpoint and illumination variations. The fourth dataset utilised in this work is the 17 places dataset \cite{sahdev2016indoor} which contains images that are captured in distinct lighting conditions, in indoor environments. For the purpose of this work, three locations were selected namely Arena, AshRoom and Corridor. Hence, the dataset consists of 420 query (\textit{day\_vme1}) and 420 reference images (\textit{night\_vme1}).
    
     To enable an image size comparison, the datasets have been resized to 224x224 pixels for the CNNs and 256x256 pixels (the closest power of 2 $-$ $2^8$x$2^8$) for HOG. Moreover, resizing the datasets also helps in comparing the time required to perform VPR for HOG with the rest of the methods, as similar image resolution will be utilised by all techniques. Following the resizing process, multiple JPEG compression ratios have been applied. Fig. \ref{averageimagesize} presents the average image size - in Kilobytes (KB) - for each of the four datasets utilised, throughout the entire spectrum of JPEG compression. We only plot the image size for the datasets resized to 256x256 pixels in Fig. \ref{averageimagesize}, as there is no major difference in size between the two types of resized datasets. It can be seen that significant size reduction can be achieved, especially when applying a high JPEG compression ratio.

     \subsection{Performance Metric}
     The accuracy \cite{9849680, tomita2021convsequential} of a VPR system is a performance metric employed to determine the effectiveness of various VPR techniques. This metric is defined as follows:
     \begin{equation}\label{eq:accuracy}
              Accuracy = \frac{No.\ of\ Correct\ Matches}{No.\ of\ Map\ Images}\;
        \end{equation}

    The accuracy of any given VPR system analyses the percentage of correctly matched images, with values in range [0,1]. Thus, higher accuracy denotes better place matching performance.

    \section{Results and Analysis}\label{results}
    This section presents the analysis of each sequence-based VPR technique on the employed test data. The sequence length that enables perfect place matching performance throughout the entire spectrum of JPEG compression is reported, utilising both uniformly (same compression ratio is utilised for both query and map images) and non-uniformly (compression ratio differs from query to map images) compressed datasets. An analysis on the amount of data transferred and the time required to perform VPR for each technique utilised is also included in this section. All experiments have been conducted on a PC equipped with an Intel Core i7-4790k CPU.

    \begin{figure*}[t]
            \centering
            \begin{tabular}{ c c c c }

                \includegraphics[width=114pt, trim=8 8 8 8, clip]{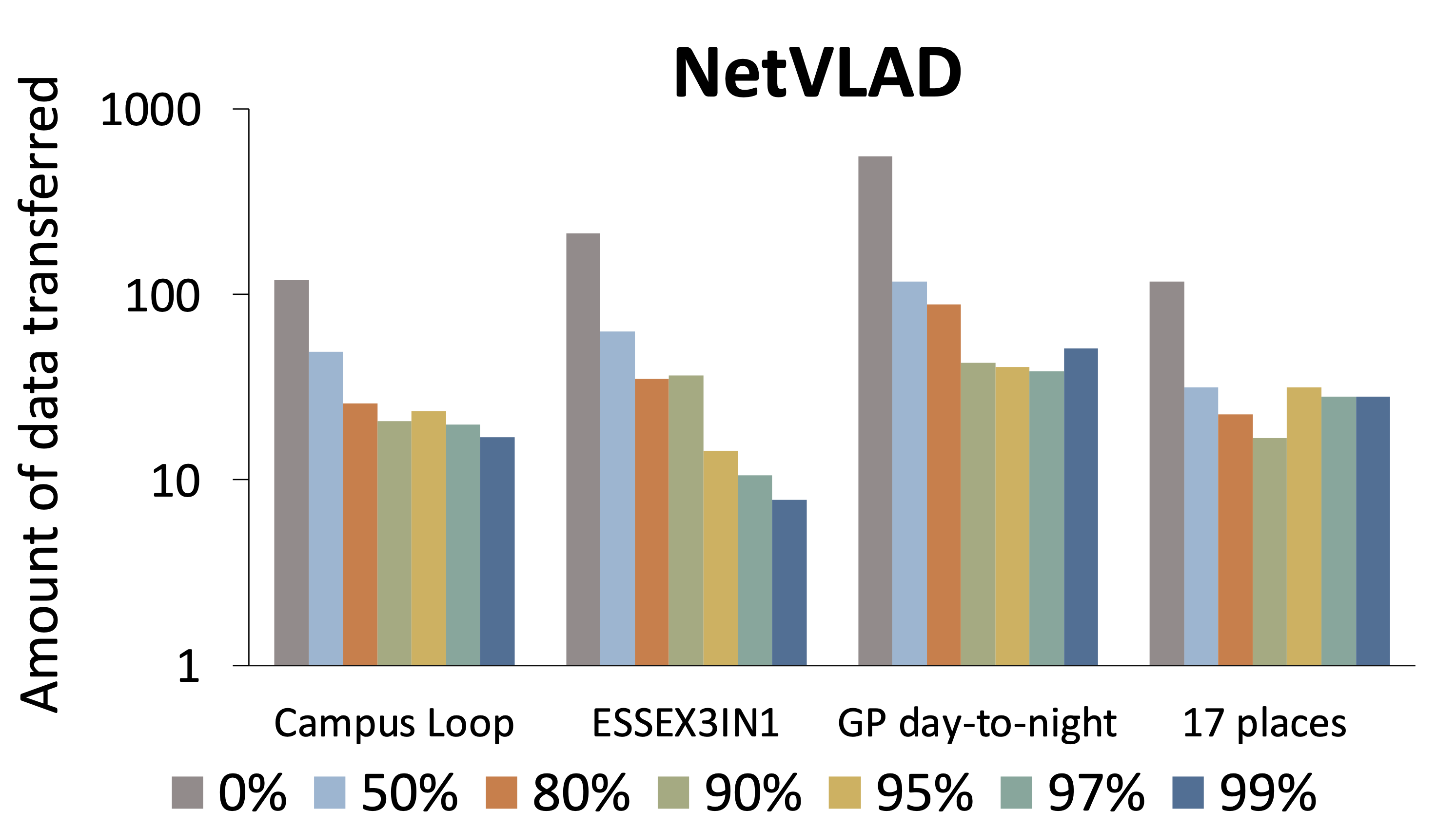} & 
                \includegraphics[width=114pt, trim=8 8 8 8, clip]{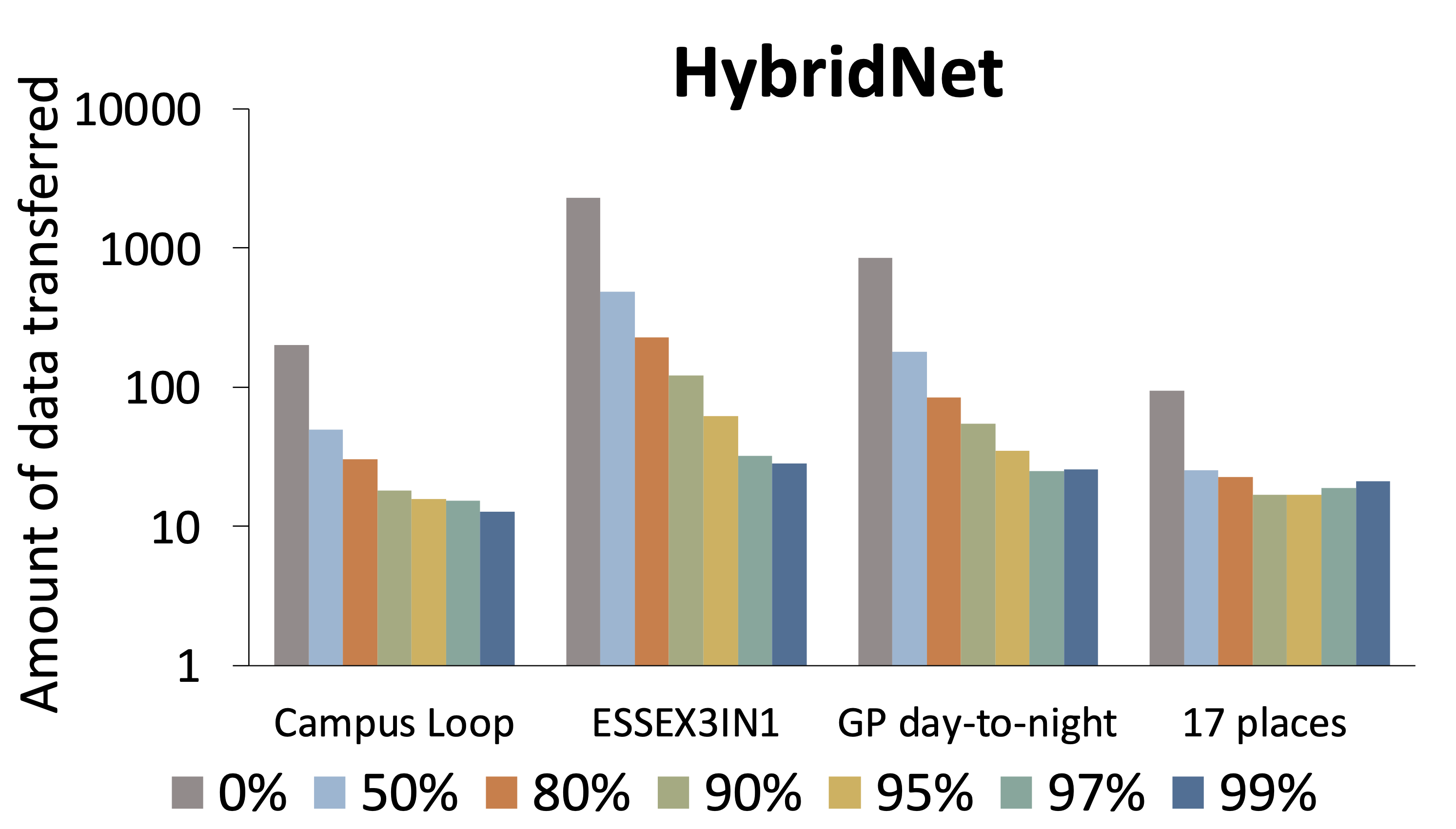} &
                \includegraphics[width=114pt, trim=8 8 8 8, clip]{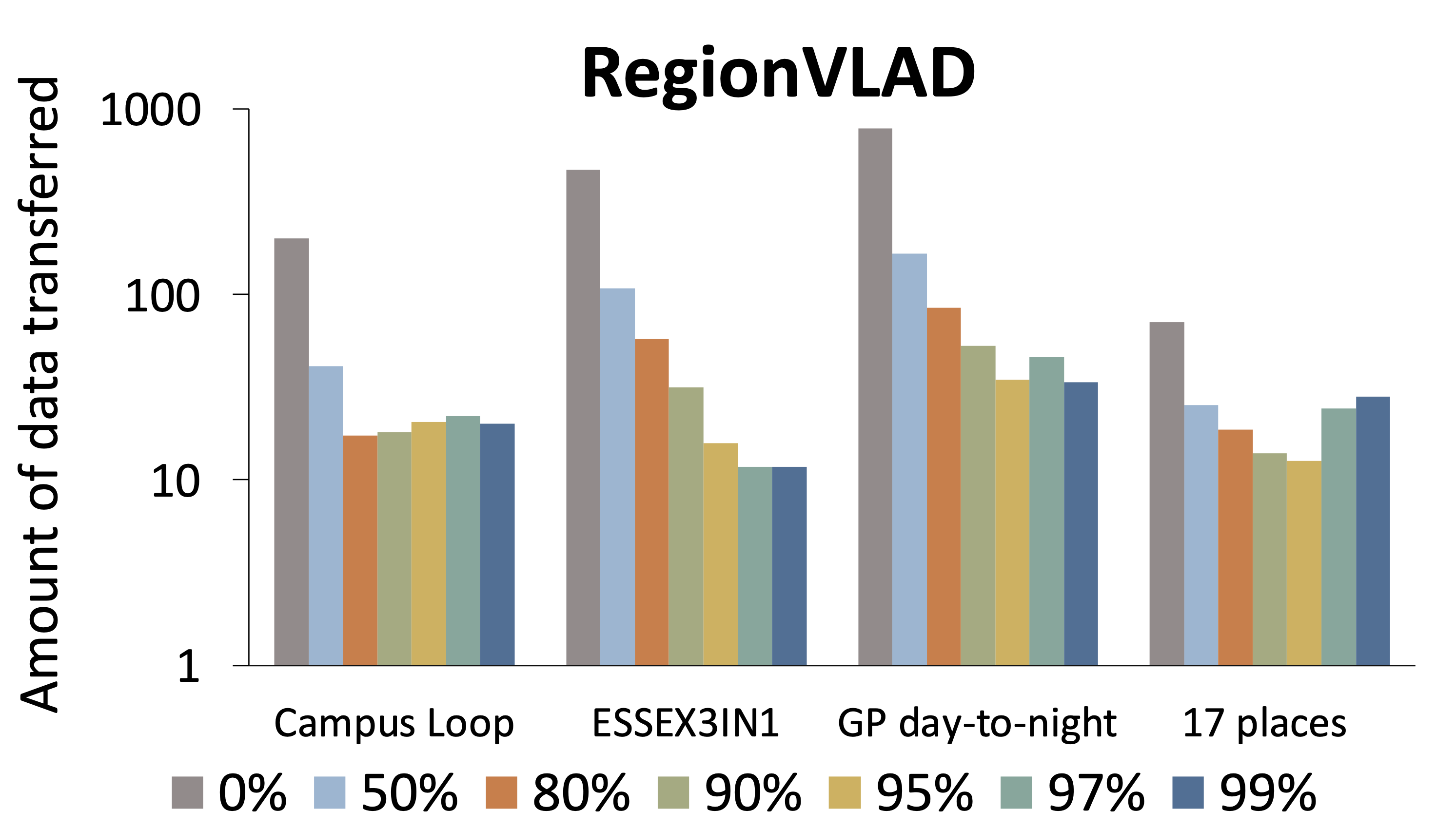} &
                \includegraphics[width=114pt, trim=8 8 8 8, clip]{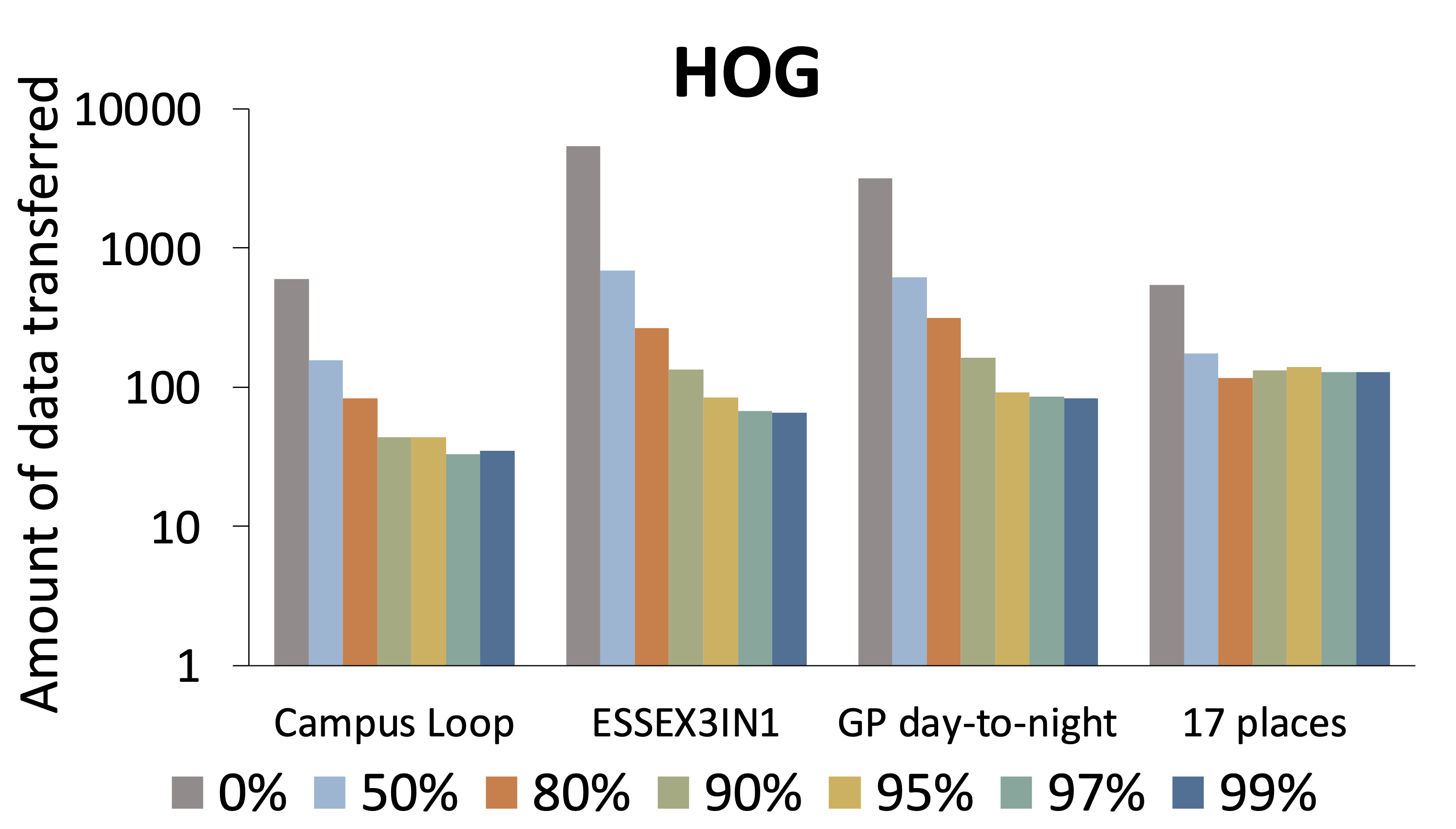}
                
            \end{tabular}
            \caption{The amount of data transferred in Kilobytes (KB) for each VPR technique and JPEG compression ratio.}
            \label{datatransferredlog}
    \end{figure*}

    \begin{figure*}[t]
             \centering
             \begin{tabular}{ c c c c }

                \includegraphics[width=114pt, trim=8 8 8 8, clip]{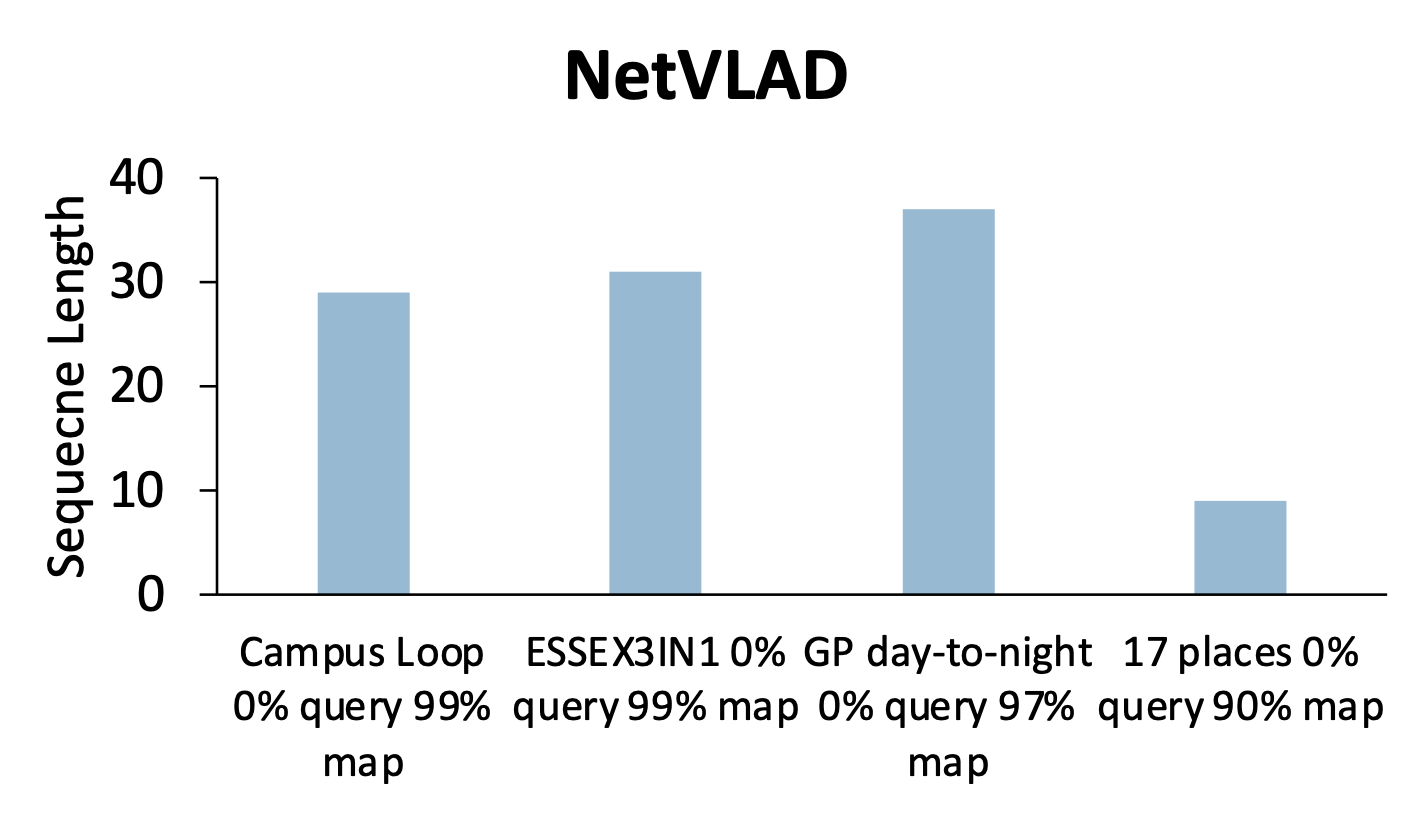} & 
                \includegraphics[width=114pt, trim=8 8 8 8, clip]{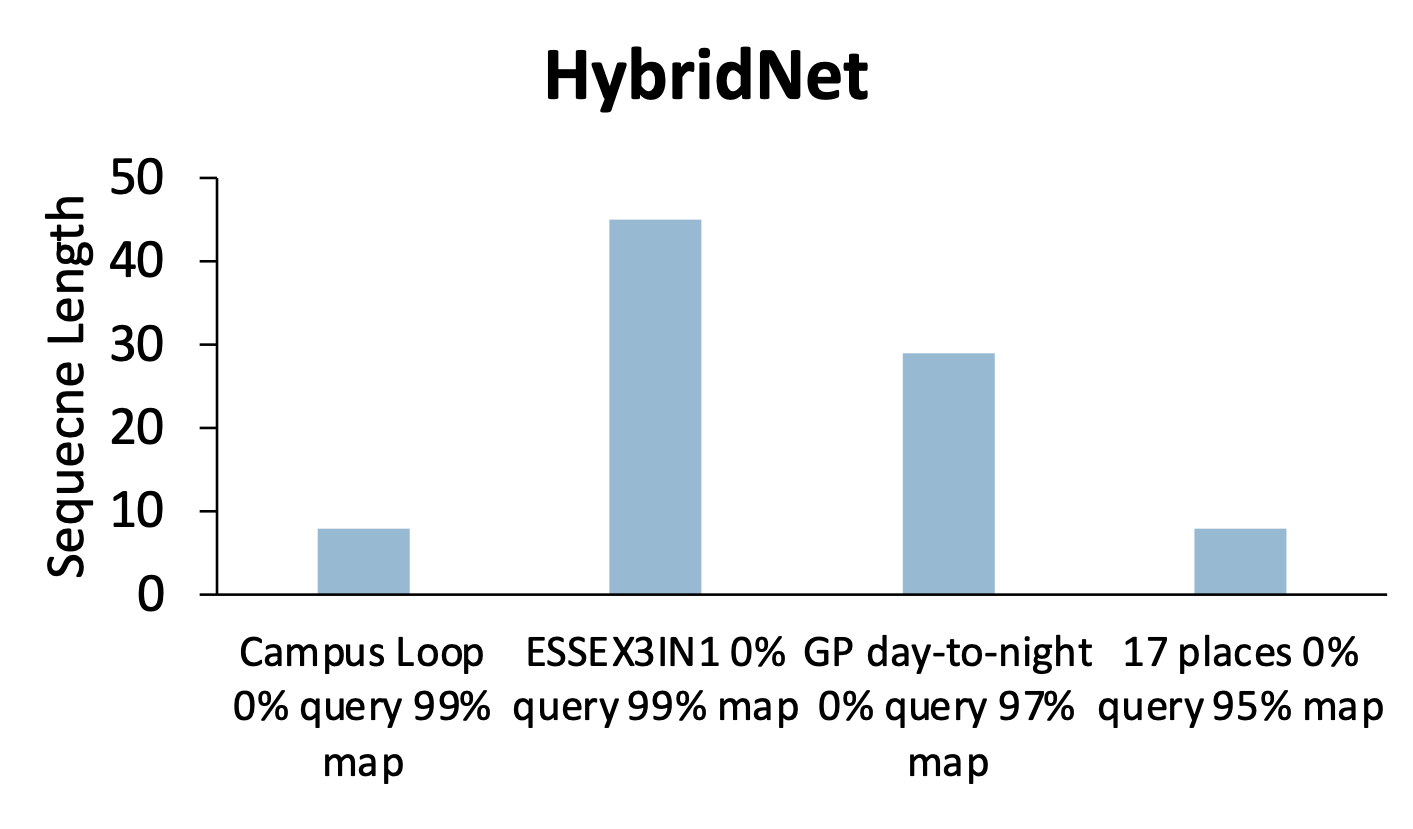} &
                \includegraphics[width=114pt, trim=8 8 8 8, clip]{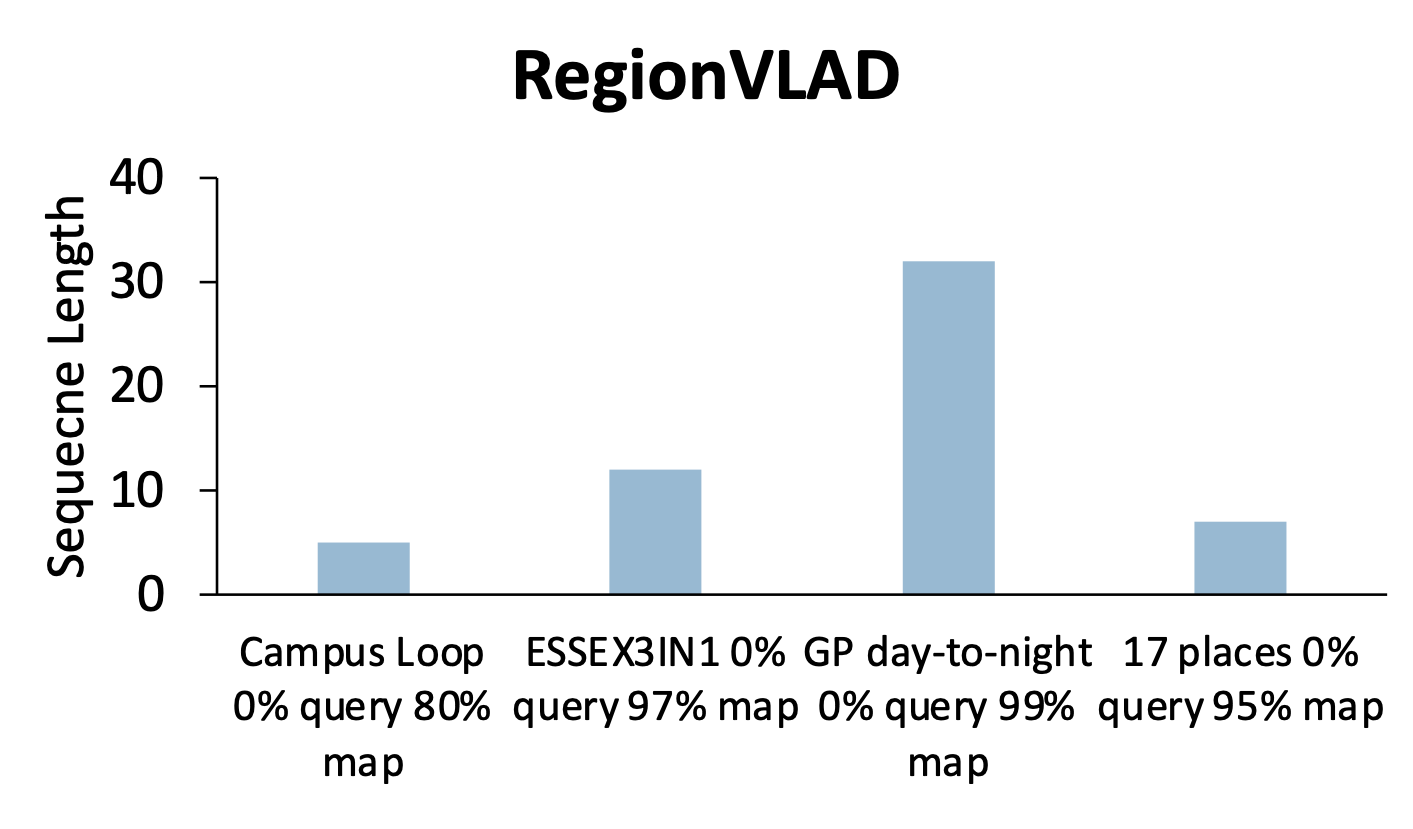} &
                \includegraphics[width=114pt, trim=8 8 8 8, clip]{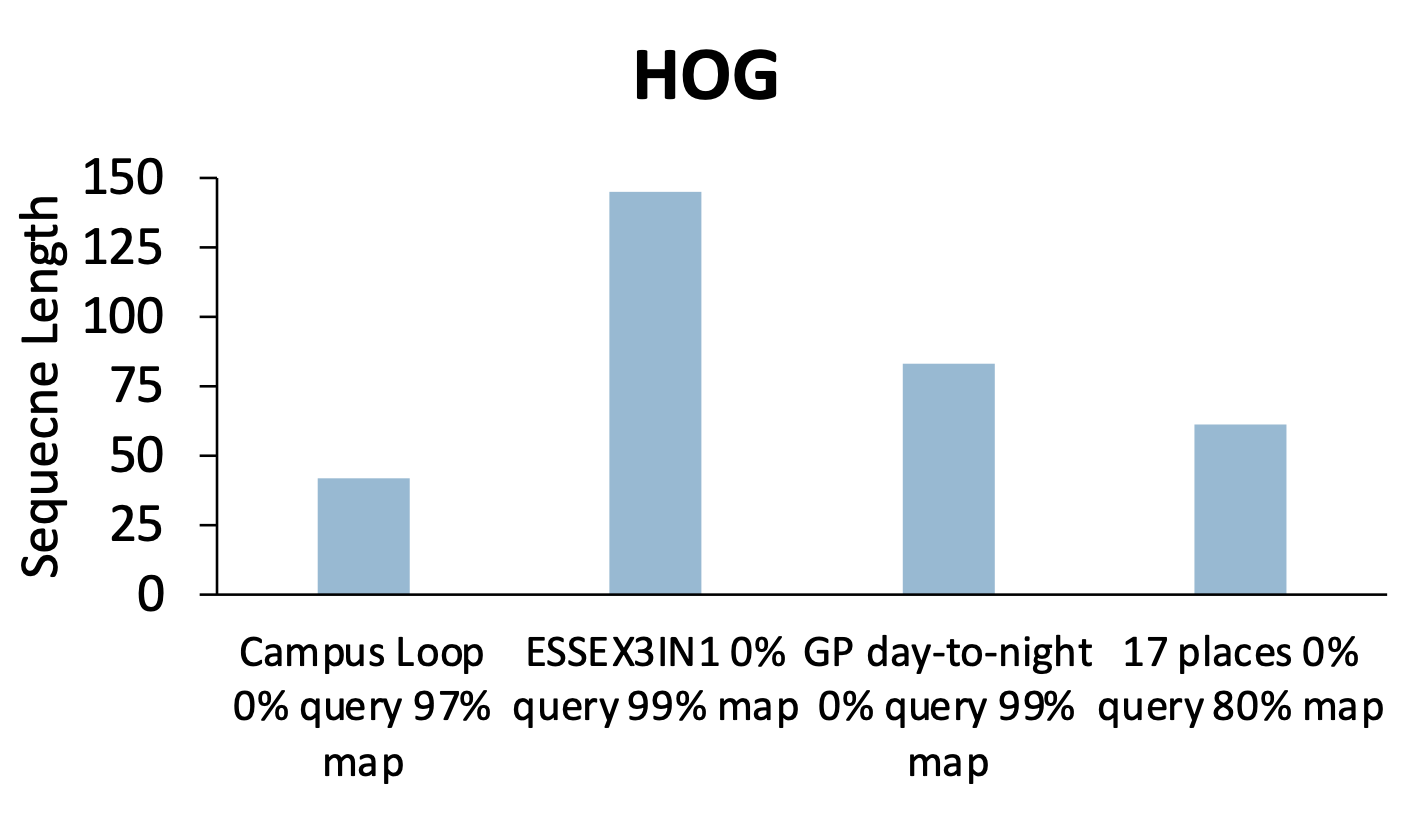} \\

                \includegraphics[width=114pt, trim=8 8 8 8, clip]{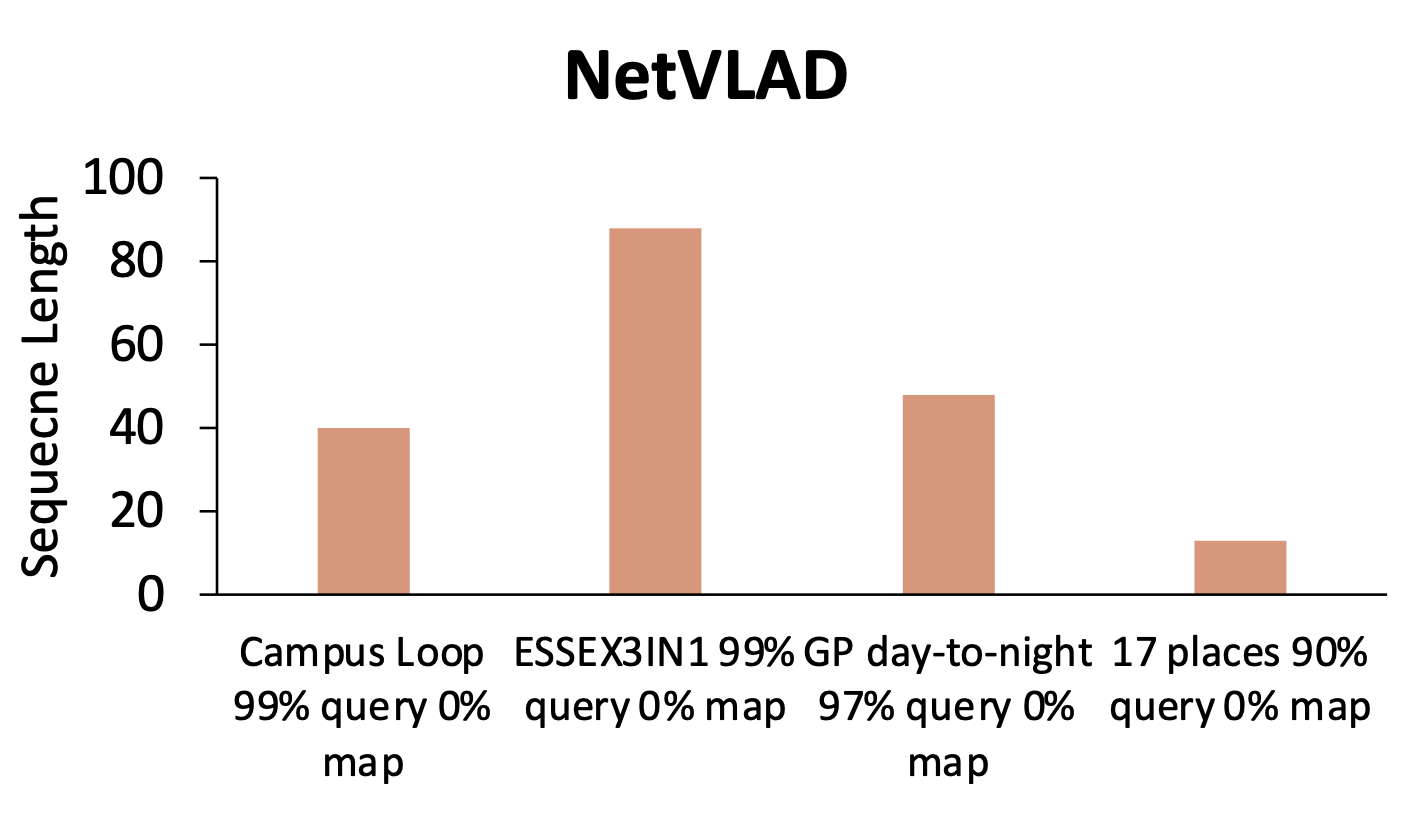} & 
                \includegraphics[width=114pt, trim=8 8 8 8, clip]{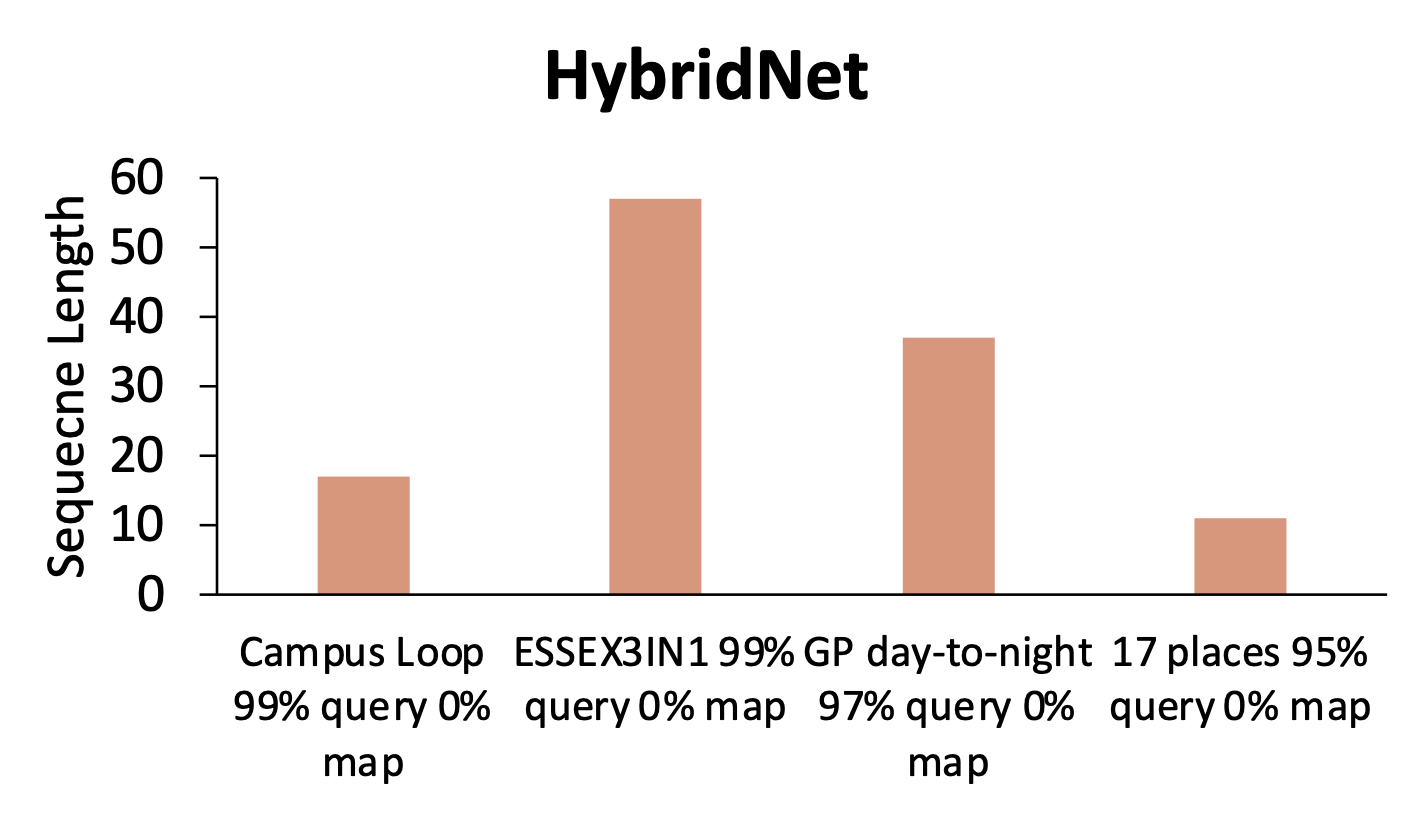} &
                \includegraphics[width=114pt, trim=8 8 8 8, clip]{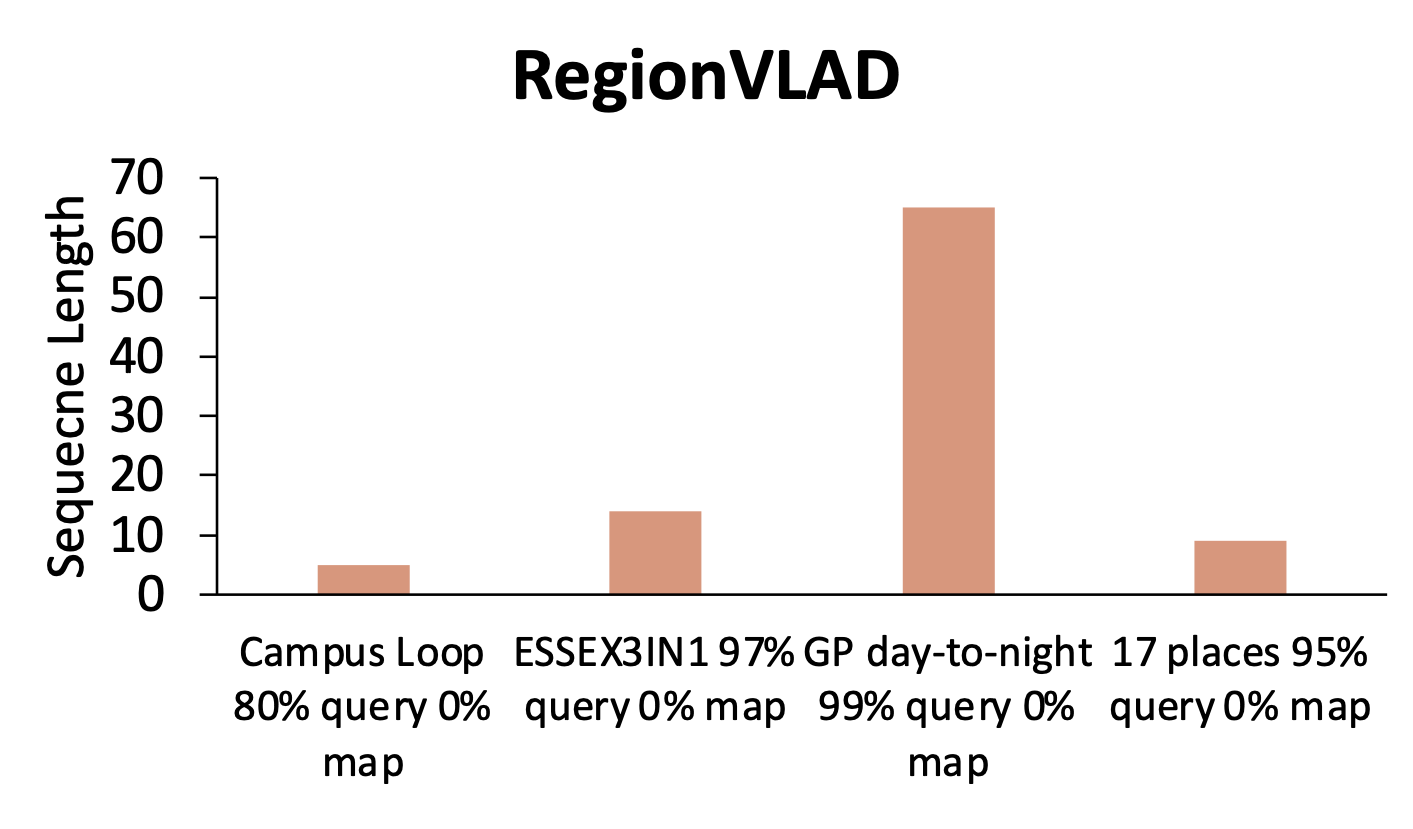} &
                \includegraphics[width=114pt, trim=8 8 8 8, clip]{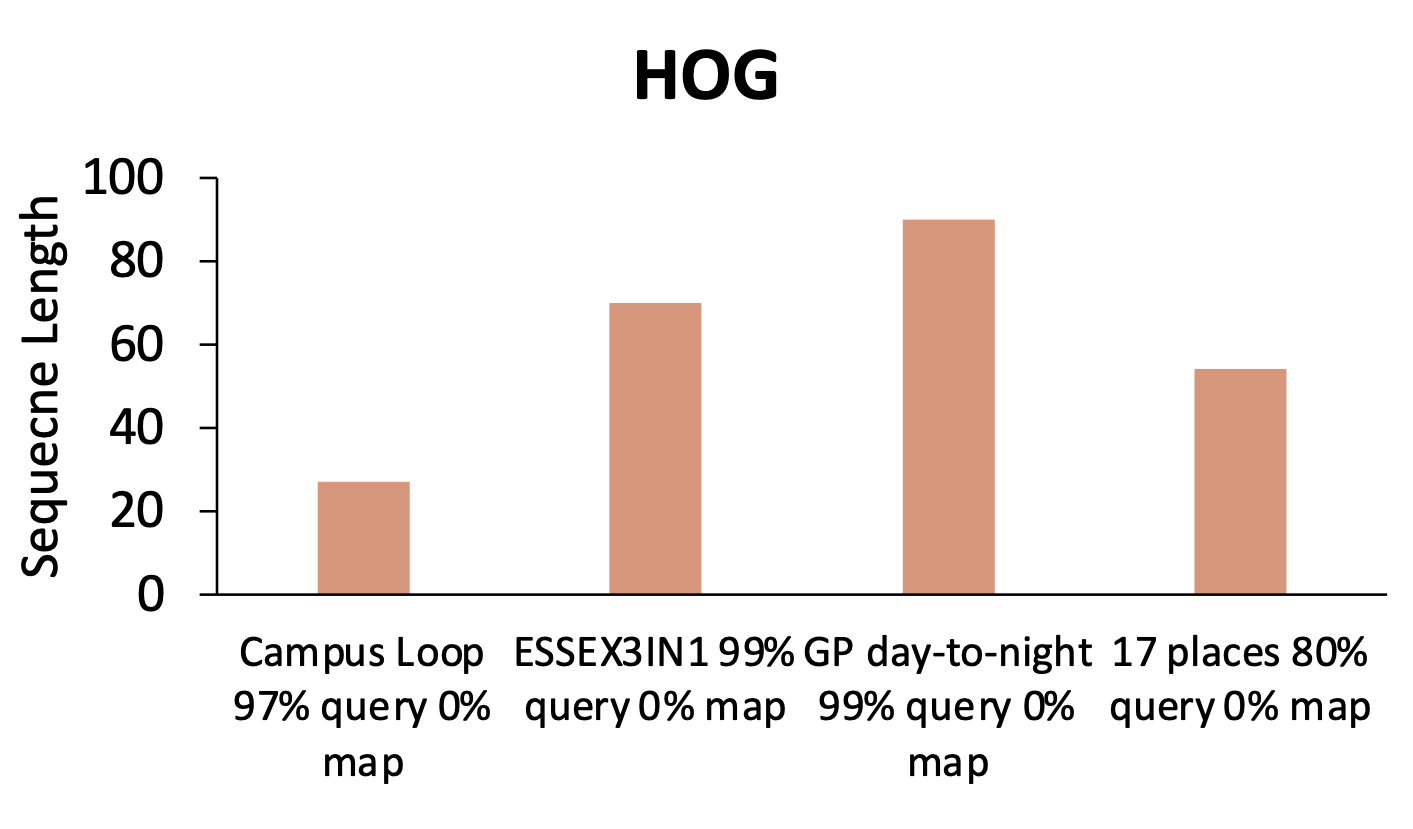}
                
            \end{tabular}
            \caption{The sequence length required to achieve maximum accuracy on non-uniformly JPEG compressed data. 
            }
            \label{non-uniform_compression}
    \end{figure*}

    \subsection{Sequence Length Impact on VPR}
    JPEG compression hinders a VPR descriptor's ability of performing successful place matching, especially in the higher spectrum of compression, as discussed in sub-section \ref{data_requirements}. However, VPR techniques designed to handle appearance changes $-$ such as HybridNet $-$ are more tolerant to high levels of JPEG compression in contrast with descriptors designed for viewpoint changes $-$ such as NetVLAD $-$ which are prone to severe loss in accuracy. To overcome the extreme loss in VPR performance resulted from introducing JPEG compression, sequence-based filtering is introduced in several VPR techniques (presented in section \ref{vprtechniques}). The details of our analysis are presented below.

    \subsection{Data Requirements for 100\% accurate VPR} \label{data_requirements}
    Fig. \ref{sequencelengthgraph} presents the sequence length \textbf{K} that facilitates each VPR technique to achieve perfect place matching performance (100\% accuracy), throughout the entire spectrum of JPEG compression. As utilising a higher level of JPEG compression usually yields a lower VPR performance, the sequence length \textbf{K} required to achieve maximum VPR accuracy is increased. This observation is highlighted in Fig. \ref{sequencelengthgraph} on Gardens Point day-to-night, where NetVLAD requires a sequence length \textbf{K} three times higher on the 99\% JPEG compressed dataset, in comparison with the uncompressed dataset. 
    In contrast with the other VPR techniques tested, HybridNet is more tolerant to JPEG compression as the sequence length required to reach 100\% accuracy is not greatly increased throughout the compression spectrum, as reported in Fig. \ref{sequencelengthgraph}. On the ESSEX3IN1 dataset, higher levels of JPEG compression improve the performance of this technique. Hence, HybridNet benefits from JPEG compression on the ESSEX3IN1 dataset, achieving maximum accuracy at 99\% JPEG compression utilising a lower sequence length \textbf{K} than on any other compression level employed. 
    
    Table \ref{table:descriptorsize} presents a comparison between the average image size taken from the ESSEX3IN1 dataset and the image descriptor size of each VPR technique. The results suggest that it would be beneficial transmitting the compressed image rather than the image descriptor. This is especially noticeable for RegionVLAD, whose descriptor size is considerably higher than any ratio of JPEG compression applied to a given image. For the remaining VPR techniques, above 50\% JPEG compression, the average image size is less than their descriptor size as shown in Table \ref{table:descriptorsize}. 



    
    
    
    
    In decentralised VPR applications where the visual data has to be shared between multiple robotic platforms, the amount of data transferred has to be carefully considered as to not hamper with the VPR process. At any given JPEG compression level, the amount of data transferred \textit{d} by a VPR technique can be calculated as follows:
    \begin{equation}\label{eq:data_amount}
            d = i_{s} * K,
    \end{equation} where \textit{$i_{s}$} is the average image size (in Kilobytes) at the given JPEG compression level, and \textbf{K} represents the sequence length that enables the VPR technique to achieve maximum place matching performance. For each VPR technique employed in this study, Fig. \ref{datatransferredlog} shows the amount of data \textit{d} required to be transferred throughout the entire spectrum of JPEG compression. A common observation is that the amount of data required for transfer \textit{d} decreases as the amount of JPEG increases. As the image's size \textit{$i_{s}$} is greatly reduced with an increase in JPEG compression (as seen in Fig. \ref{averageimagesize}), a longer sequence length would not always result in a larger amount of data transfer \textit{d}, as observed in Fig. \ref{datatransferredlog}.
    The results presented above suggest that it is beneficial to use a higher JPEG compression rate paired with a longer sequence length \textbf{K}, as it allows the same levels of VPR performance to be achieved at decreased bandwidth requirements.
    
            
            
            
            

   \begin{table}
  
  \centering
  \caption{Descriptor sizes compared to the average image size of ESSEX3IN1 at several compression levels.}
    \large{
    \begin{adjustbox}{width=\columnwidth,center}
    \begin{tabular}{|c|c|c|c|c|c|c|c|}
    \hline
    \textbf{VPR} & \textbf{Descriptor} & \multicolumn{6}{c|}{\textbf{Descriptor-Image Size Ratio [\%]}} \bigstrut\\
\cline{3-8}    \textbf{Technique} & \textbf{Size [KB]} & \textbf{0\%} & \textbf{50\%} & \textbf{80\%} & \textbf{90\%} & \textbf{95\%} & \textbf{97\%} \bigstrut\\
    \hline
    \hline
    NetVLAD & 16    & 309.3 & 65.5  & 32    & 18.3  & 9.9   & 7.3 \bigstrut\\
    \hline
    HybridNet & 30    & 165   & 34.9  & 17.1  & 9.7   & 5.3   & 3.9 \bigstrut\\
    \hline
    RegionVLAD & 384   & 12.9  & 2.7   & 1.3   & 0.8   & 0.4   & 0.3 \bigstrut\\
    \hline
    HOG   & 31.6  & 156.6 & 33.2  & 16.2  & 9.2   & 5     & 3.7 \bigstrut\\
    \hline
    \end{tabular}%
    \end{adjustbox}
    }
  \label{table:descriptorsize}%
\end{table}%

\begin{figure*}[t]
            \centering
            \begin{tabular}{ c c c c }

                \includegraphics[width=114pt, trim=8 8 8 8, clip]{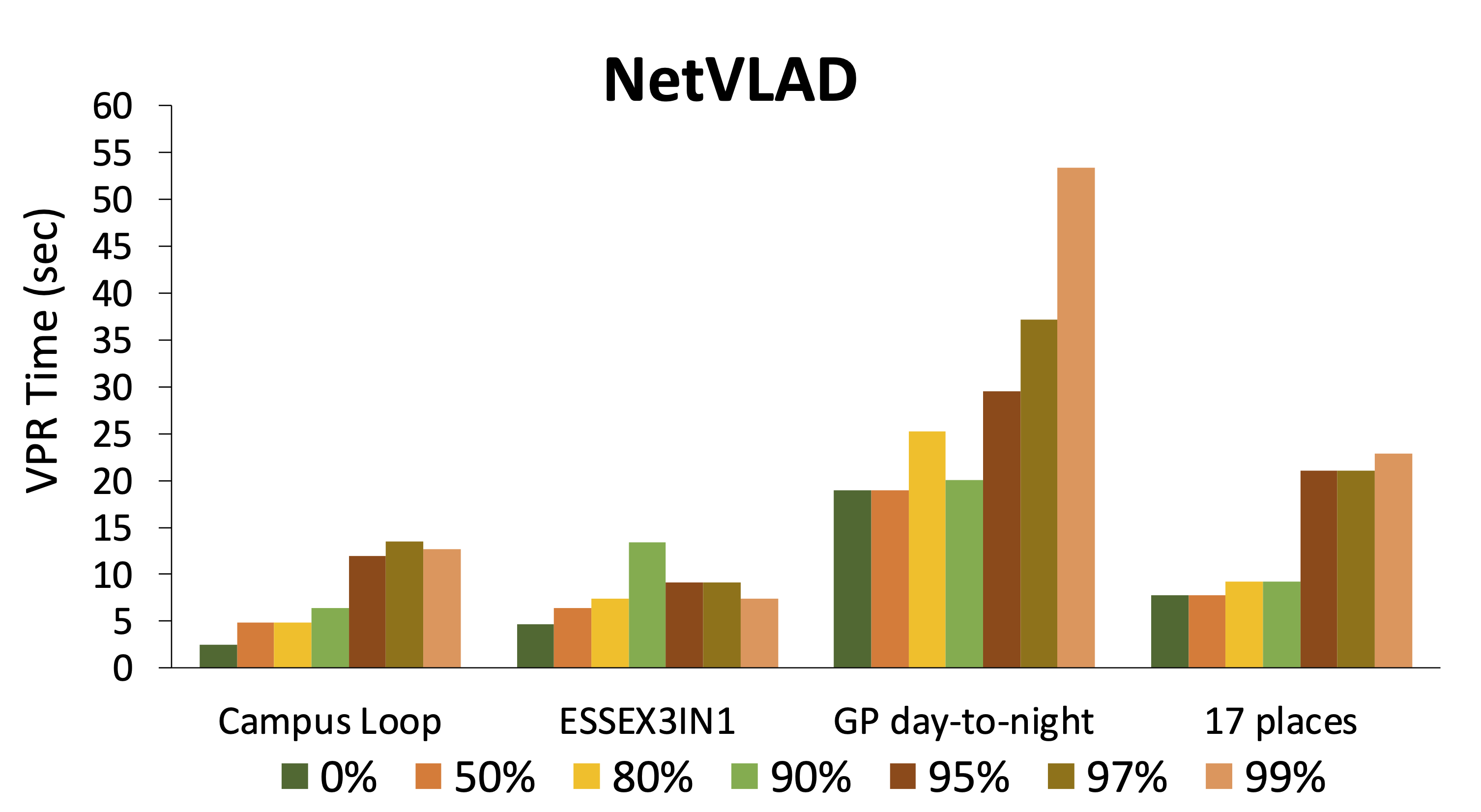} & 
                \includegraphics[width=114pt, trim=8 8 8 8, clip]{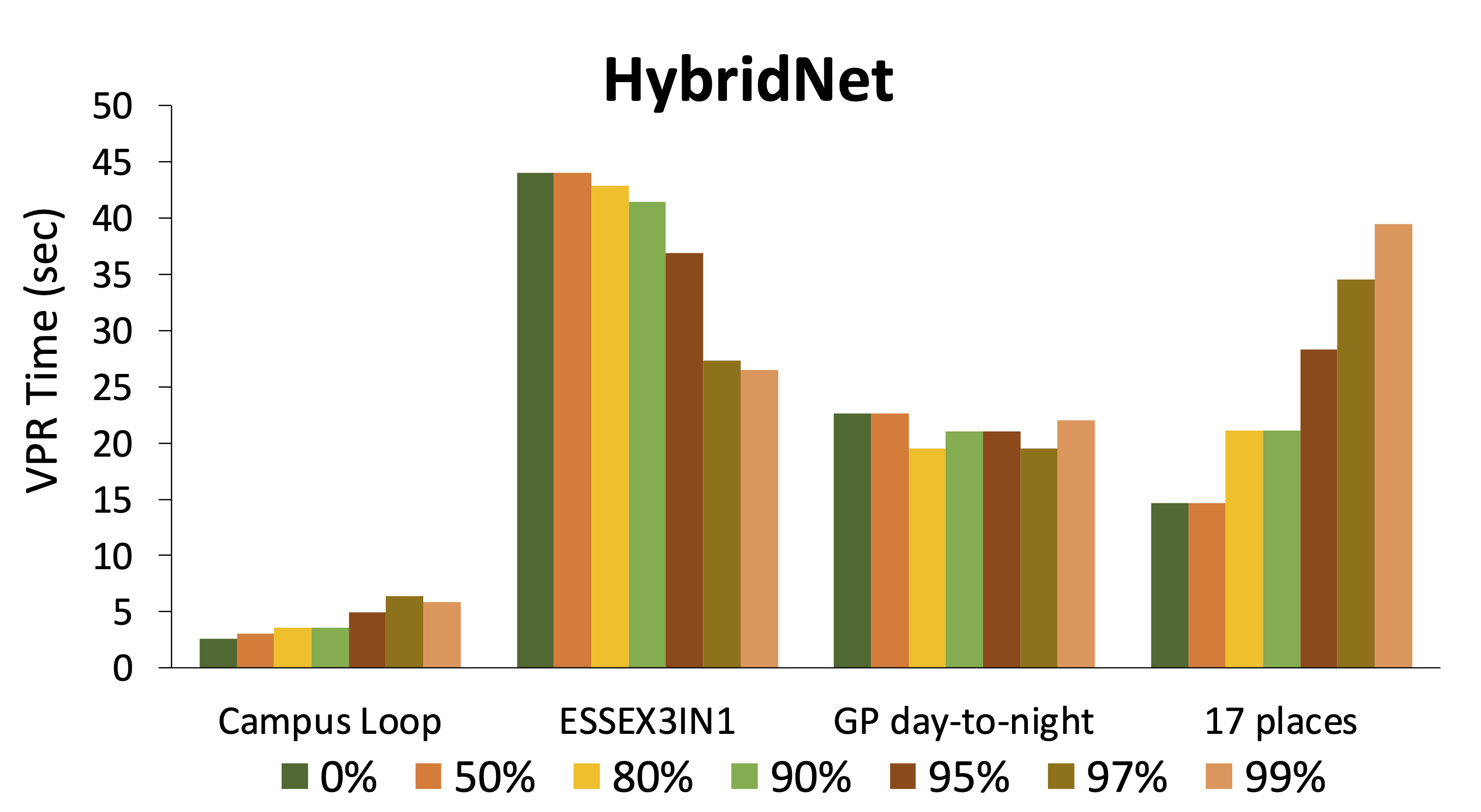} &
                \includegraphics[width=114pt, trim=8 8 8 8, clip]{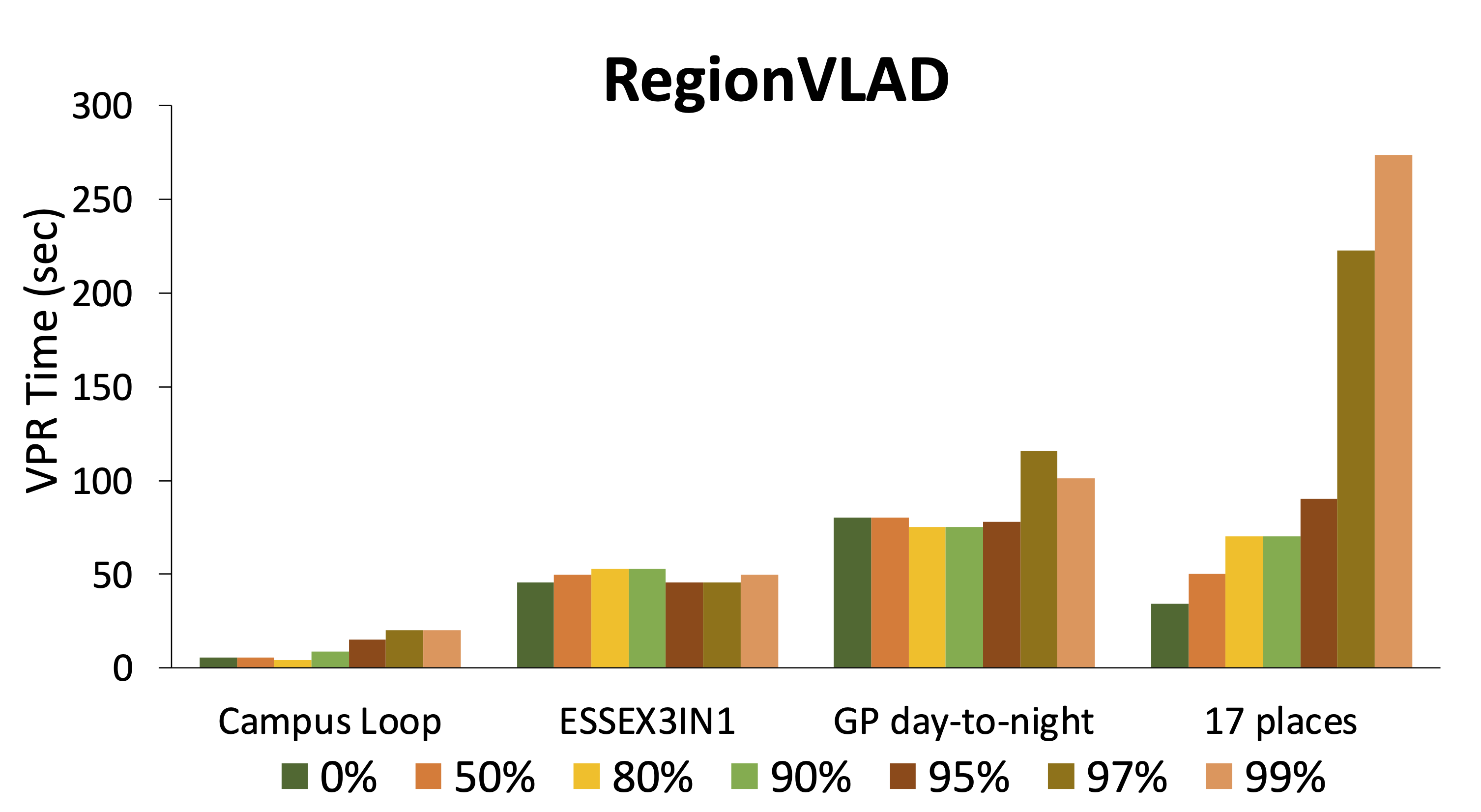} &
                \includegraphics[width=114pt, trim=8 8 8 8, clip]{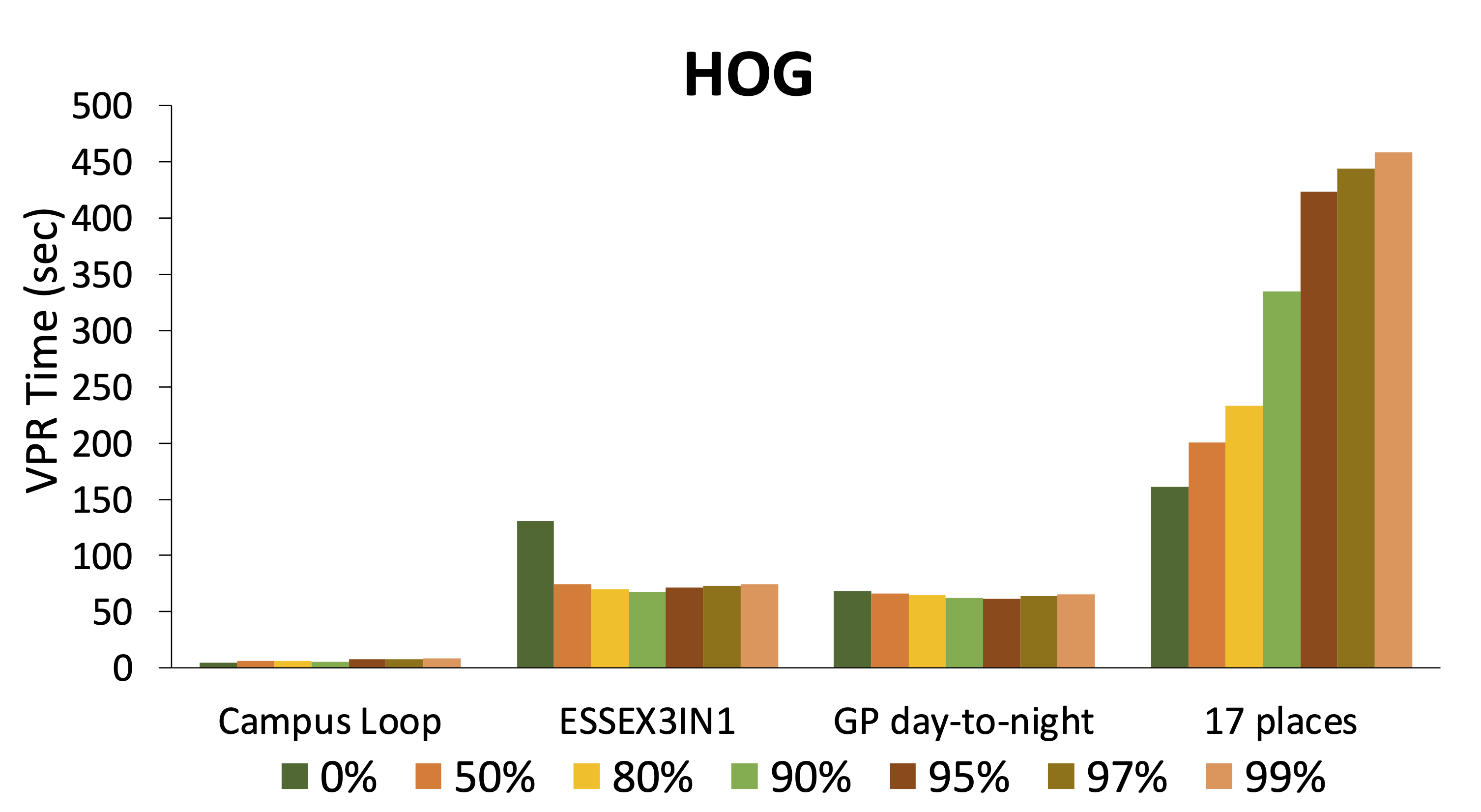}
                
            \end{tabular}
            \caption{$t_{VPR}$ for every VPR technique on each dataset and JPEG compression amount specified in Fig. \ref{sequencelengthgraph}}
            \label{VPRtime_uniform}
            \end{figure*}

             \begin{figure*}[t]
             \centering
             \begin{tabular}{ c c c c }

                \includegraphics[width=114pt, trim=8 8 8 8, clip]{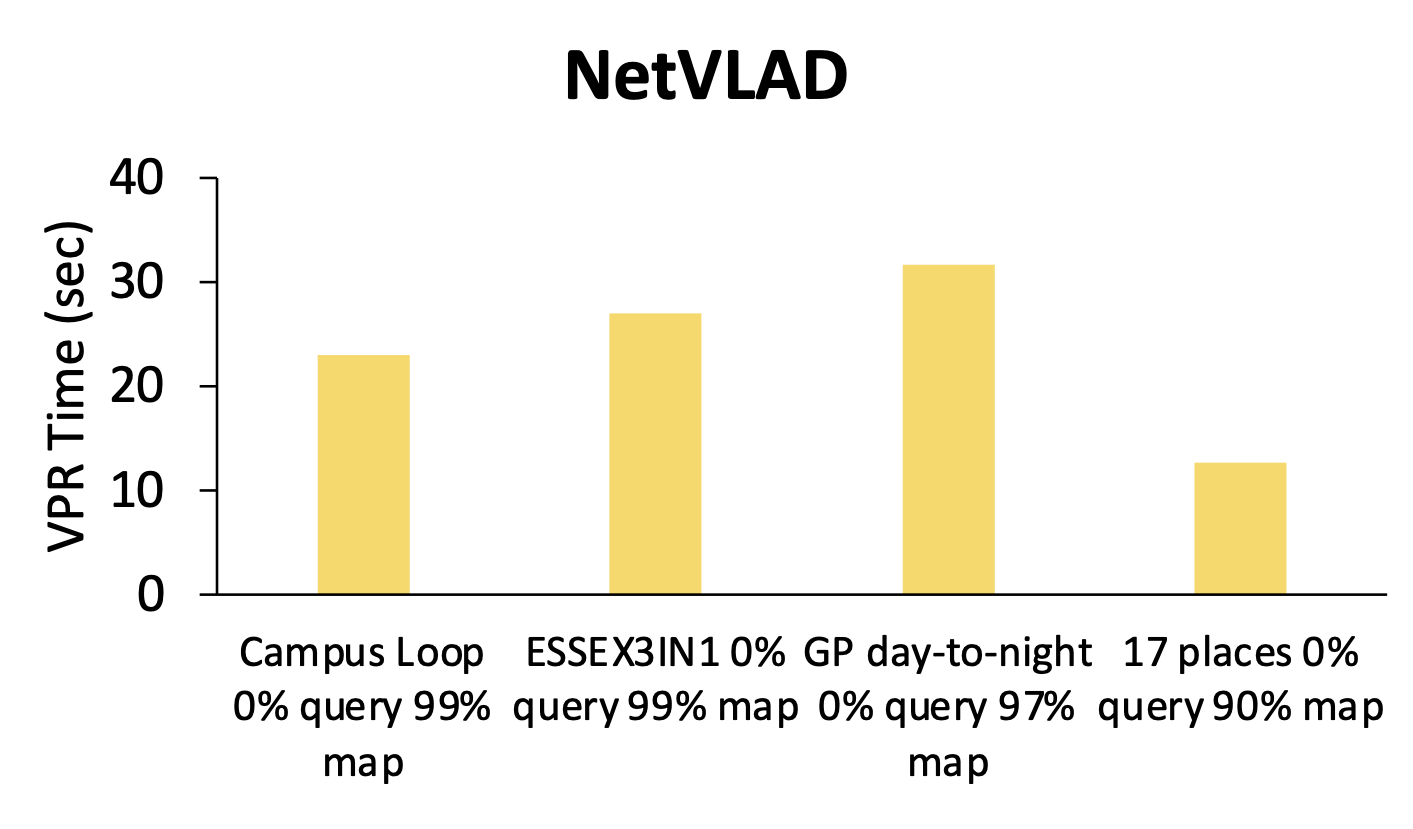} & 
                \includegraphics[width=114pt, trim=8 8 8 8, clip]{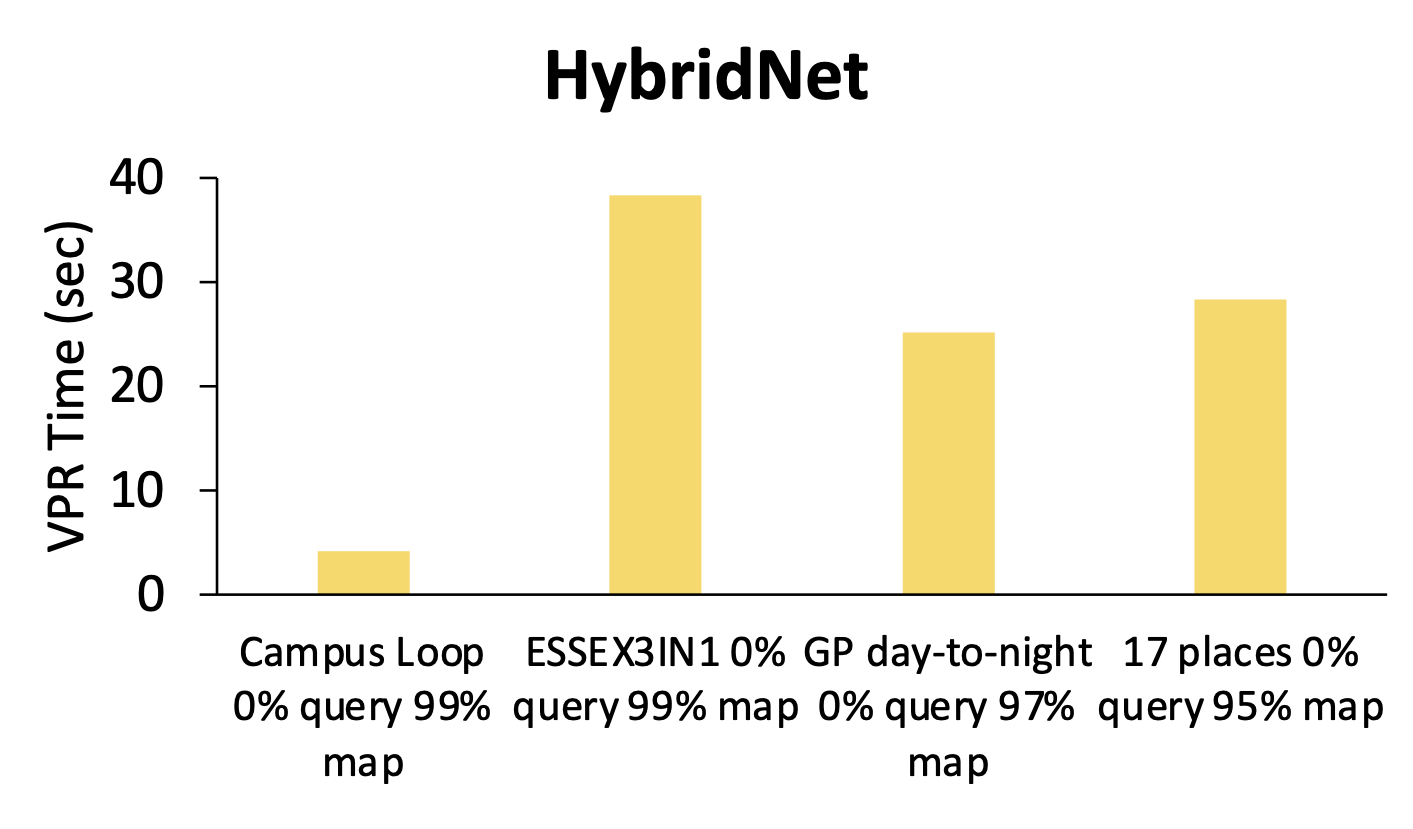} &
                \includegraphics[width=114pt, trim=8 8 8 8, clip]{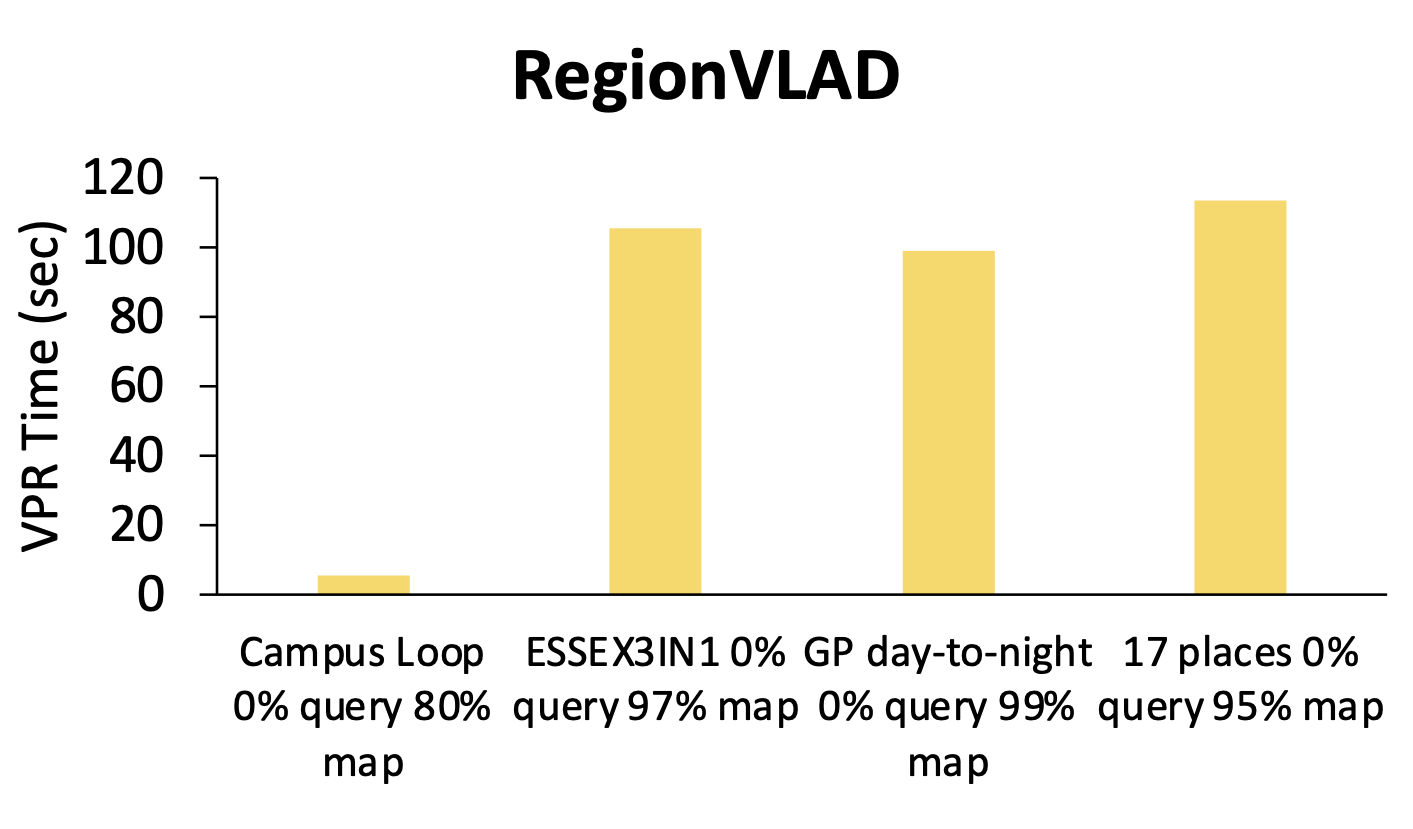} &
                \includegraphics[width=114pt, trim=8 8 8 8, clip]{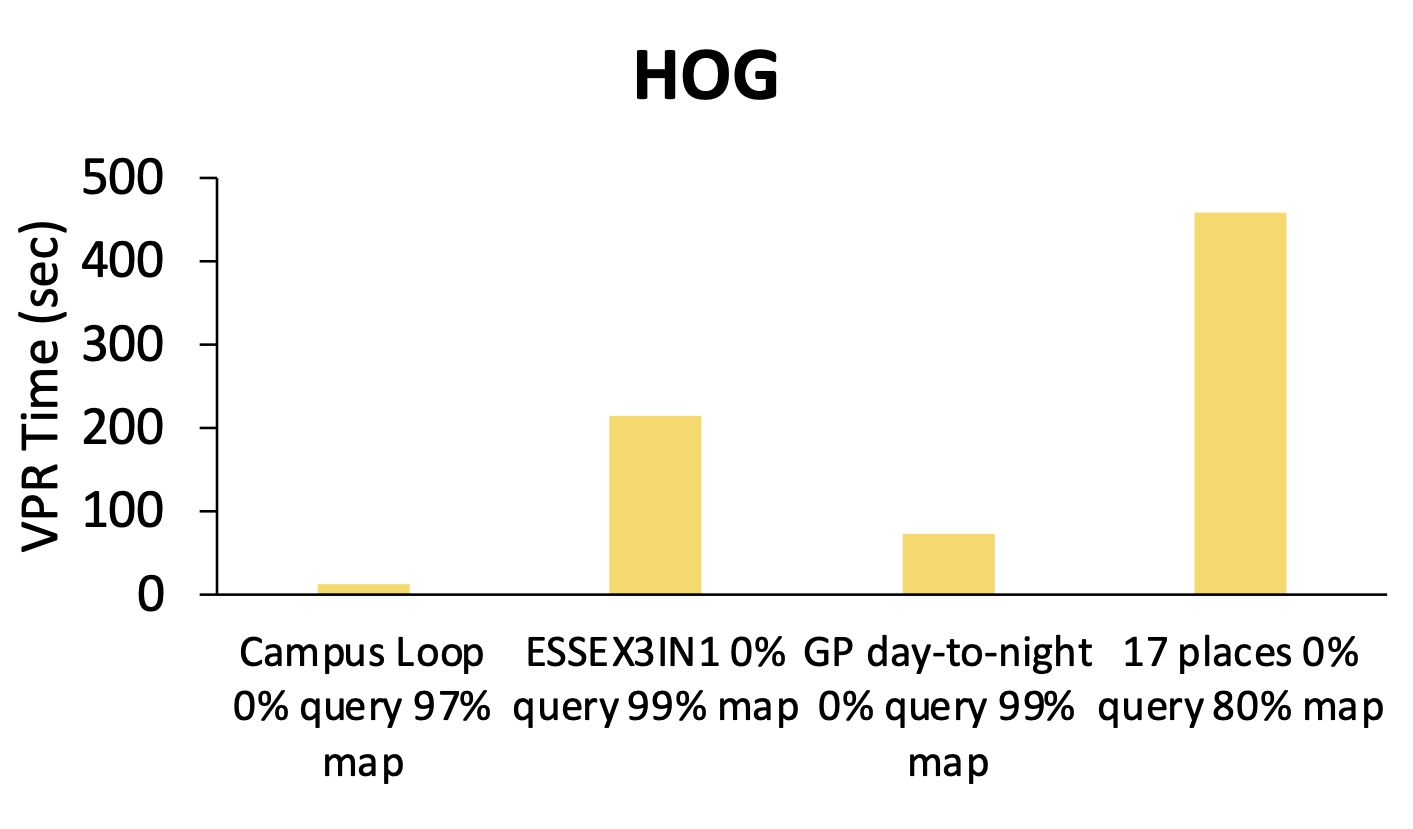} \\

                \includegraphics[width=114pt, trim=8 8 8 8, clip]{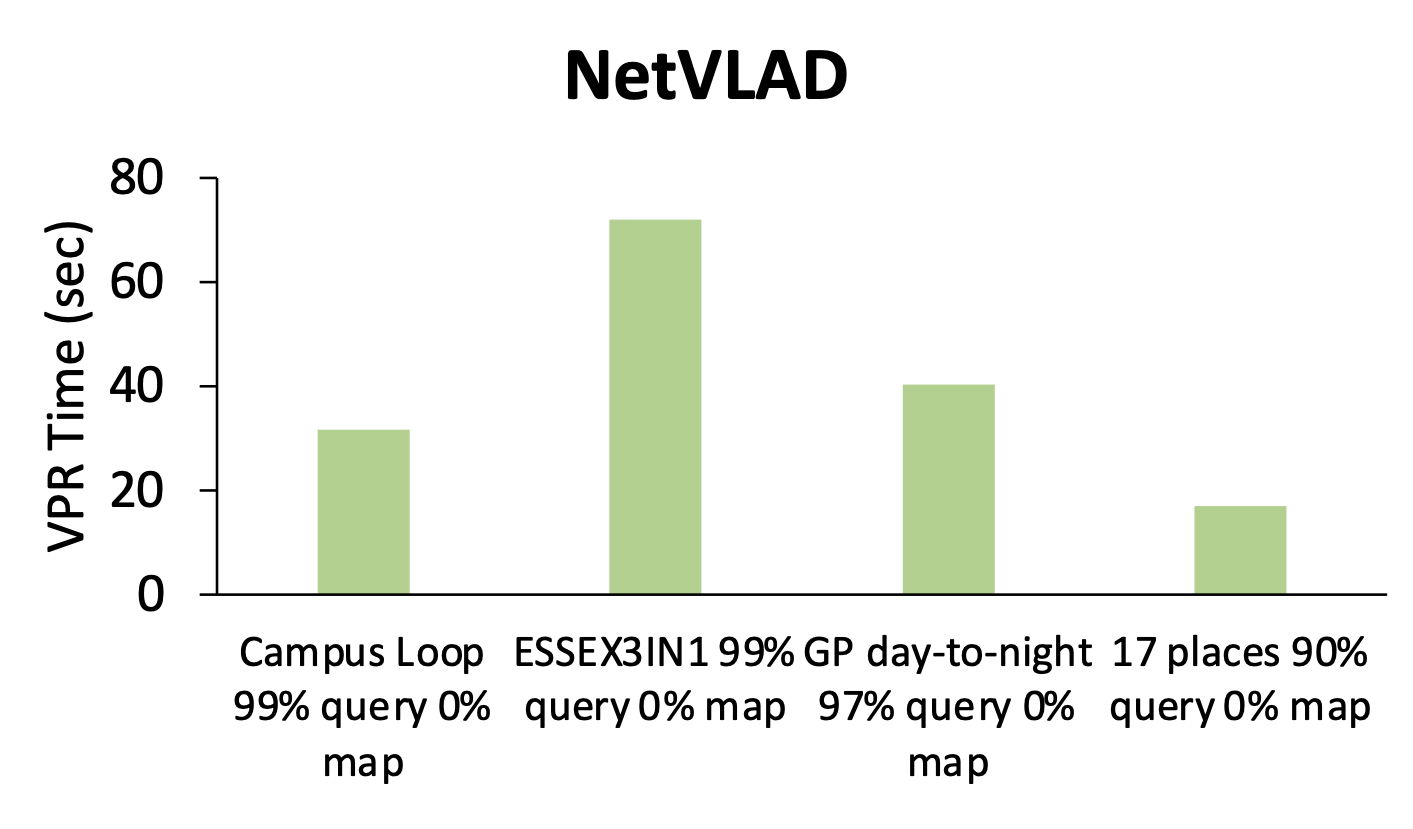} & 
                \includegraphics[width=114pt, trim=8 8 8 8, clip]{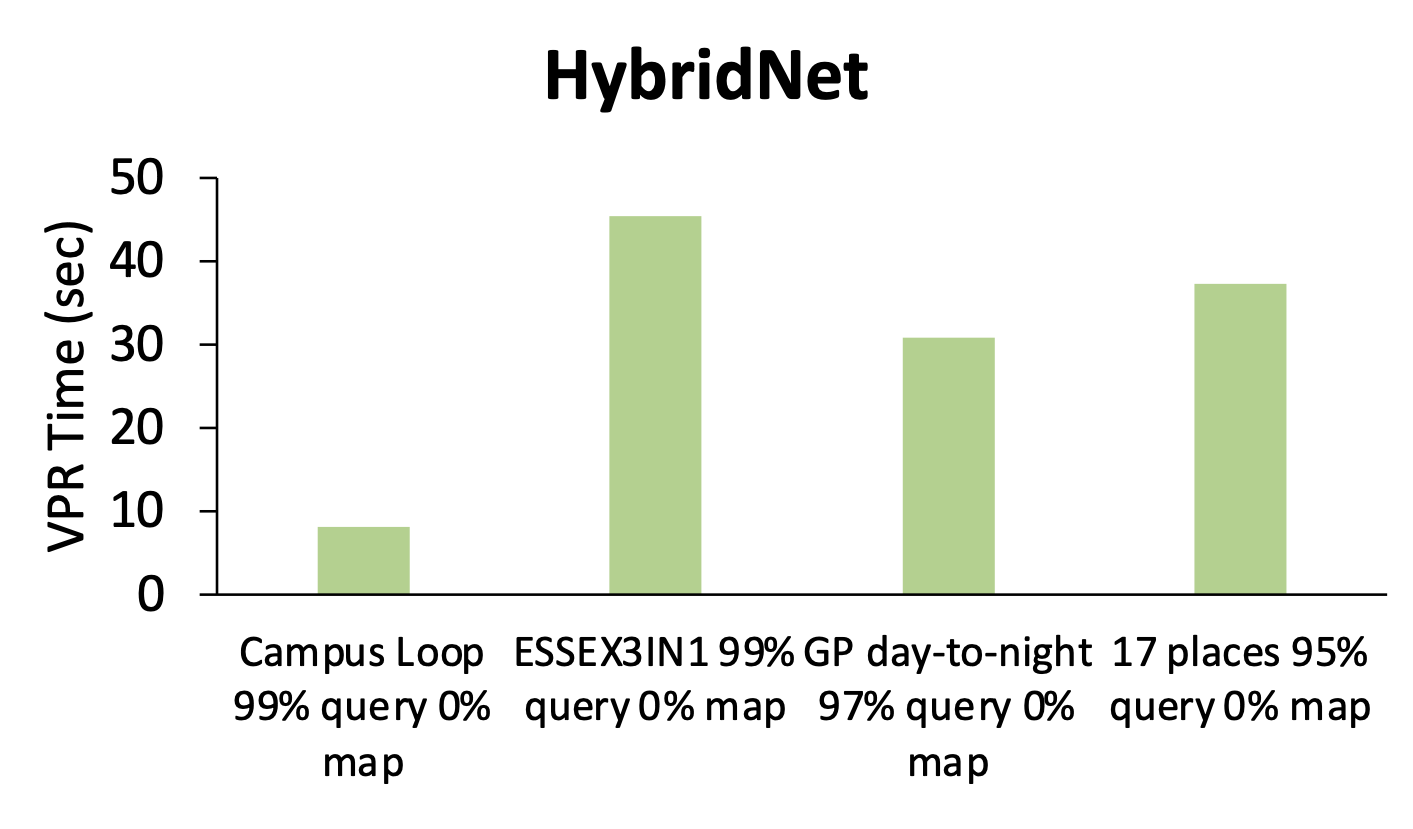} &
                \includegraphics[width=114pt, trim=8 8 8 8, clip]{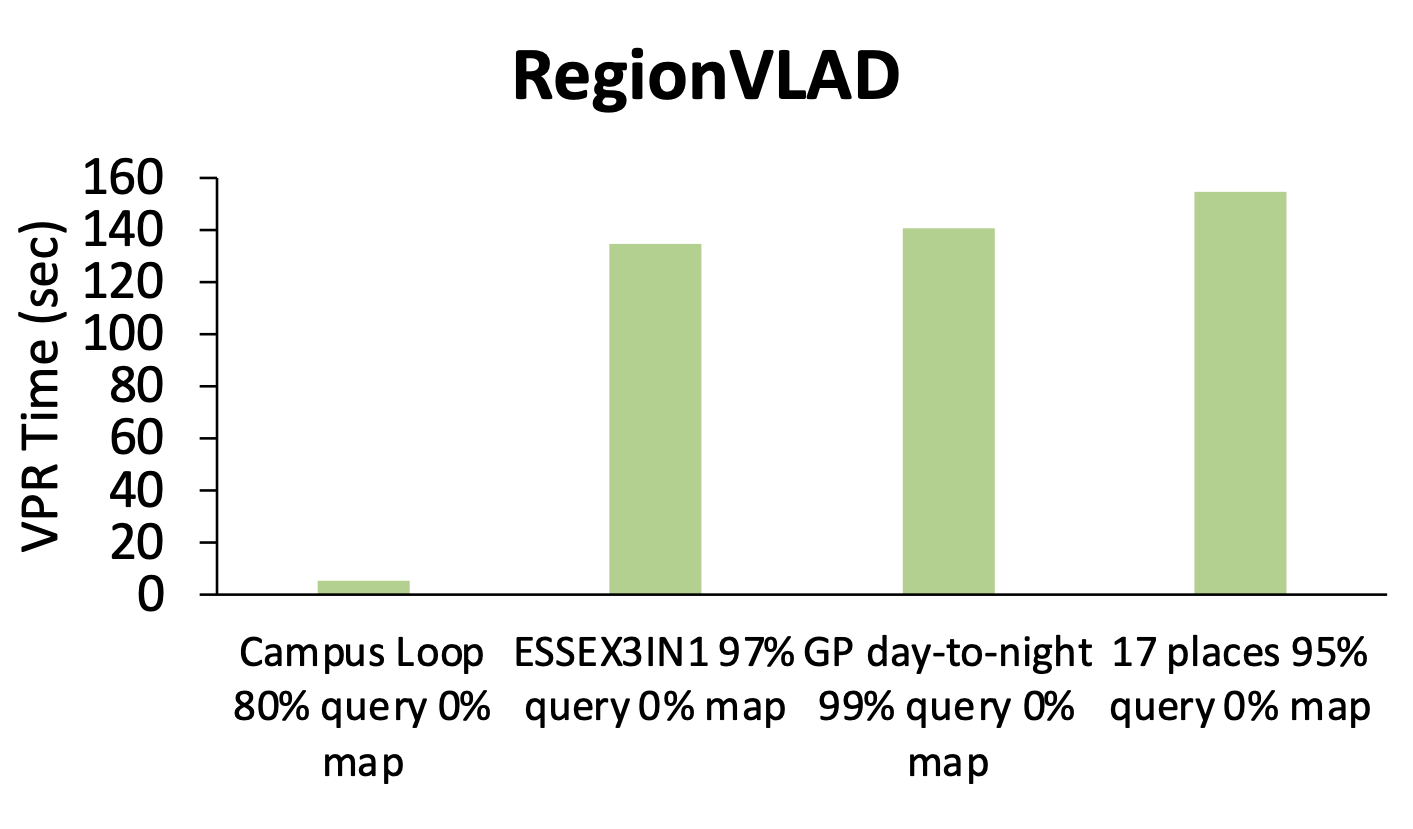} &
                \includegraphics[width=114pt, trim=8 8 8 8, clip]{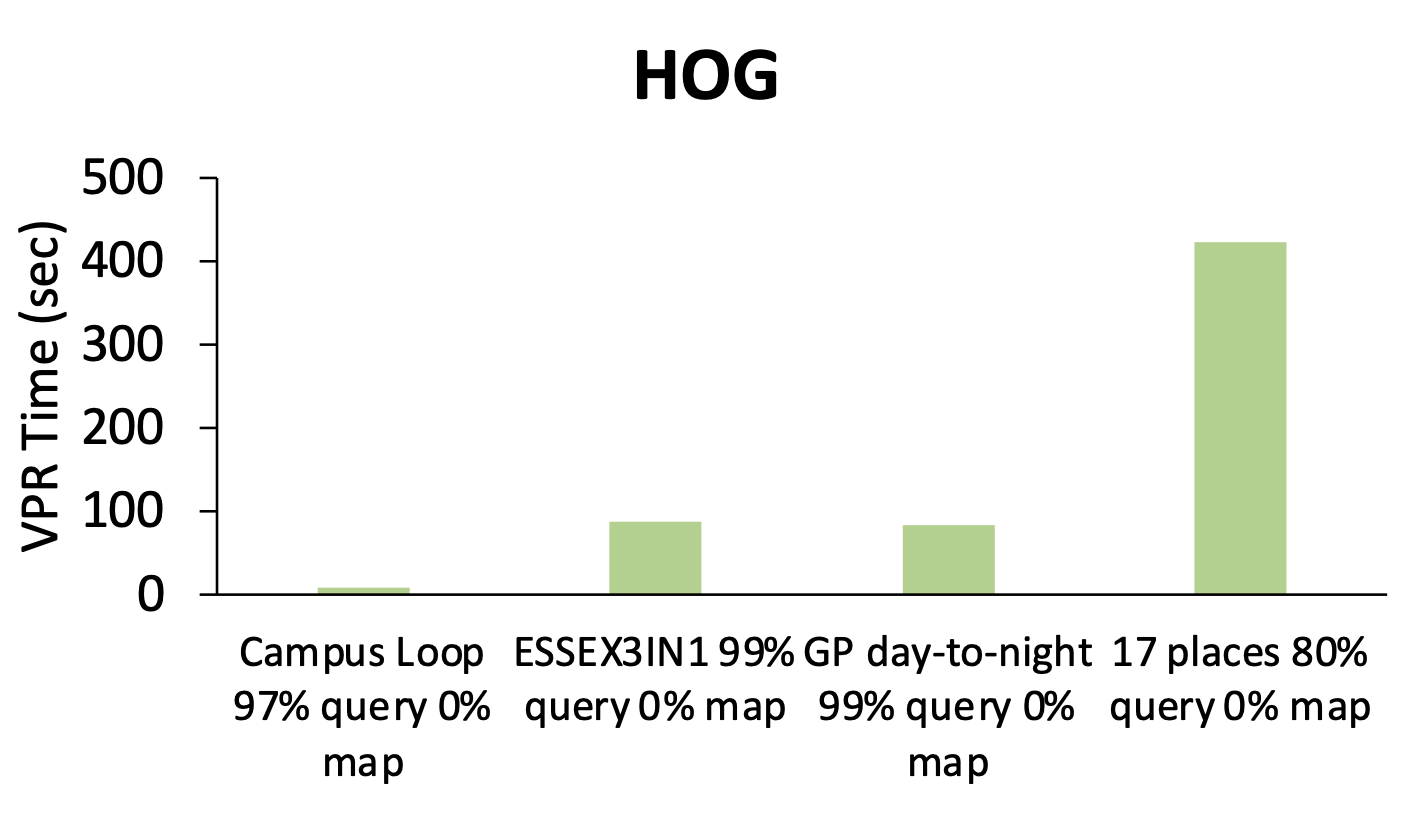}
                
            \end{tabular}
            \caption{$t_{VPR}$ of each VPR technique on non-uniformly JPEG compressed data specified in Fig. \ref{non-uniform_compression}. 
            }
            \label{VPRtime_non-uniform}
            \end{figure*}

\begin{table*}[t]
        
        \centering
        \caption{Total VPR time, in scenarios where the ESSEX3IN1 dataset is both uniformly and non-uniformly JPEG compressed. We show that more efficient VPR can be performed by utilising an uniformly JPEG compressed dataset with a shorter value of K. }
        \begin{tabular}{ |c|c|c|c|c|c|c|}
    \hline
     & \multicolumn{6}{c|}{\textbf{NetVLAD} ($t_{e}$ = 0.77 sec)} \\\cline{2-7}
     \textbf{JPEG Compression} & \textit{K}  & \textit{$t_{m}$} & \textit{$t'_{e}$} & \textit{$t_{VPR}$} & \textit{$t'_{c}$} & \textit{$t_{total}$} \\  
    
    \hline
    99\% query, 99\% map & 8 & 1.205 & 6.16 & 7.365 & 0.0102 & \textbf{7.375} \\ 
    \hline
    0\% query, 99\% map & 31 & 3.1 & 23.87 & 26.97 & - & 26.97\\
    \hline
    99\% query, 0\% map & 88 & 4.1 & 67.76 & 71.86 & - & 71.86 \\
    \hline
   & \multicolumn{6}{c|}{\textbf{HybridNet} ($t_{e}$ = 0.36 sec)} \\\cline{2-7}
     \textbf{JPEG Compression} & \textit{K}  & \textit{$t_{m}$} & \textit{$t'_{e}$} & \textit{$t_{VPR}$} & \textit{$t'_{c}$} & \textit{$t_{total}$} \\  
    
    \hline
    99\% query, 99\% map & 29 & 16.05 & 10.44 & 26.49 & 0.037 & \textbf{26.52} \\ 
    \hline
    0\% query, 99\% map & 45 & 22.14 & 16.2 & 38.34 & - & 38.34\\
    \hline
    99\% query, 0\% map & 57 & 24.95 & 20.52 & 45.47 & - & 45.47 \\
   \hline
   & \multicolumn{6}{c|}{\textbf{RegionVLAD} ($t_{e}$ = 0.424 sec)} \\\cline{2-7}
     \textbf{JPEG Compression} & \textit{K}  & \textit{$t_{m}$} & \textit{$t'_{e}$} & \textit{$t_{VPR}$} & \textit{$t'_{c}$} & \textit{$t_{total}$} \\  
    
    \hline
    97\% query, 97\% map & 11 & 40.96 & 4.664 & 45.624 & 0.0142 & \textbf{45.638} \\ 
    \hline
    0\% query, 97\% map & 30 & 92.78 & 12.72 & 105.5 & - & 105.5\\
    \hline
    97\% query, 0\% map & 45 & 115.78 & 19.08 & 134.86 & - & 134.86 \\
   \hline
   & \multicolumn{6}{c|}{\textbf{HOG} ($t_{e}$ = 0.0043 sec)} \\\cline{2-7}
     \textbf{JPEG Compression} & \textit{K}  & \textit{$t_{m}$} & \textit{$t'_{e}$} & \textit{$t_{VPR}$} & \textit{$t'_{c}$} & \textit{$t_{total}$} \\  
    
    \hline
    99\% query, 99\% map & 61 & 73.94 & 0.262 & 74.2 & 0.0778 & \textbf{74.277} \\ 
    \hline
    0\% query, 99\% map & 145 & 214.09 & 0.623 & 214.71 & - & 214.71\\
    \hline
    99\% query, 0\% map & 70 & 88.3 & 0.301 & 88.6 & - & 88.6 \\
   \hline
    \end{tabular}
    \label{table:time_comparison}
    \end{table*}  

    \subsection{Non-Uniform Compression Ratios} \label{non-uniform_compression_ratios}
    To facilitate VPR applications where the limited bandwidth may disrupt the localisation process, the query and the map images may have different ratios of JPEG compression applied. Due to the discrepancy between the amount of JPEG compression applied to the query and reference images, the VPR performance may be drastically reduced. In Fig. \ref{non-uniform_compression}, the sequence length required by each VPR technique to reach maximum accuracy on the non-uniformly JPEG compressed datasets is assessed. For each technique, we utilise the JPEG compression ratio that would result in the minimal amount of data transferred. The results presented in Fig. \ref{non-uniform_compression}
    show that each of the tested methods tend to perform worse on non-uniformly JPEG compressed data, requiring longer sequence lengths \textbf{K} to achieve maximum performance than on uniformly compressed data (refer to Fig. \ref{sequencelengthgraph}). Moreover, Fig. \ref{non-uniform_compression} also shows that by utilising a JPEG compressed query image and uncompressed map, most VPR techniques have a decrease in performance over the scenario where only the map is compressed. 

    \subsection{Analysis on the Time Required to Perform VPR}
        
        Fig. \ref{VPRtime_uniform} presents the VPR time of each technique for the sequence lengths \textbf{K} that are required to reach maximum place matching performance on each dataset and JPEG compression ratio (\textbf{K} values are presented in Fig. \ref{sequencelengthgraph}). In the case of a given sequence-based technique, the feature encoding time $t'_{e}$ can be obtained by multiplying the feature encoding time of the single-image-based technique $t_{e}$ (presented in Table \ref{table:time_comparison}) with the number of images in a sequence \textbf{K}, as follows:
        \begin{equation}\label{eq:encodingtime}
            t_{e}' = t_e * K
            \end{equation} 
        
        To obtain the VPR time $t_{VPR}$, the matching time $t_m$ is summed with the feature encoding time $t'_{e}$ as follows:
        \begin{equation}\label{eq:vprtime}
            t_{VPR} = t_m + t_e'
            \end{equation}  
            

        As previously mentioned in sub-section \ref{non-uniform_compression_ratios} and shown in Fig. \ref{non-uniform_compression}, the sequence length \textbf{K} required to achieve 100\% accuracy when non-uniformly JPEG compressed datasets are employed is considerably higher in most cases than on uniformly compressed data. 
        The results presented in both Fig. \ref{VPRtime_uniform} and Fig. \ref{VPRtime_non-uniform} clearly show that the sequence length \textbf{K} has a direct impact on $t_{VPR}$. As a result, for any given JPEG compression ratio, a VPR technique can drastically have its $t_{VPR}$ increased if a longer sequence length \textbf{K} is employed.
    
\begin{figure}
            \centering
            \begin{tabular}{ c }

                \includegraphics[width=230pt, trim=8 8 8 8, clip]{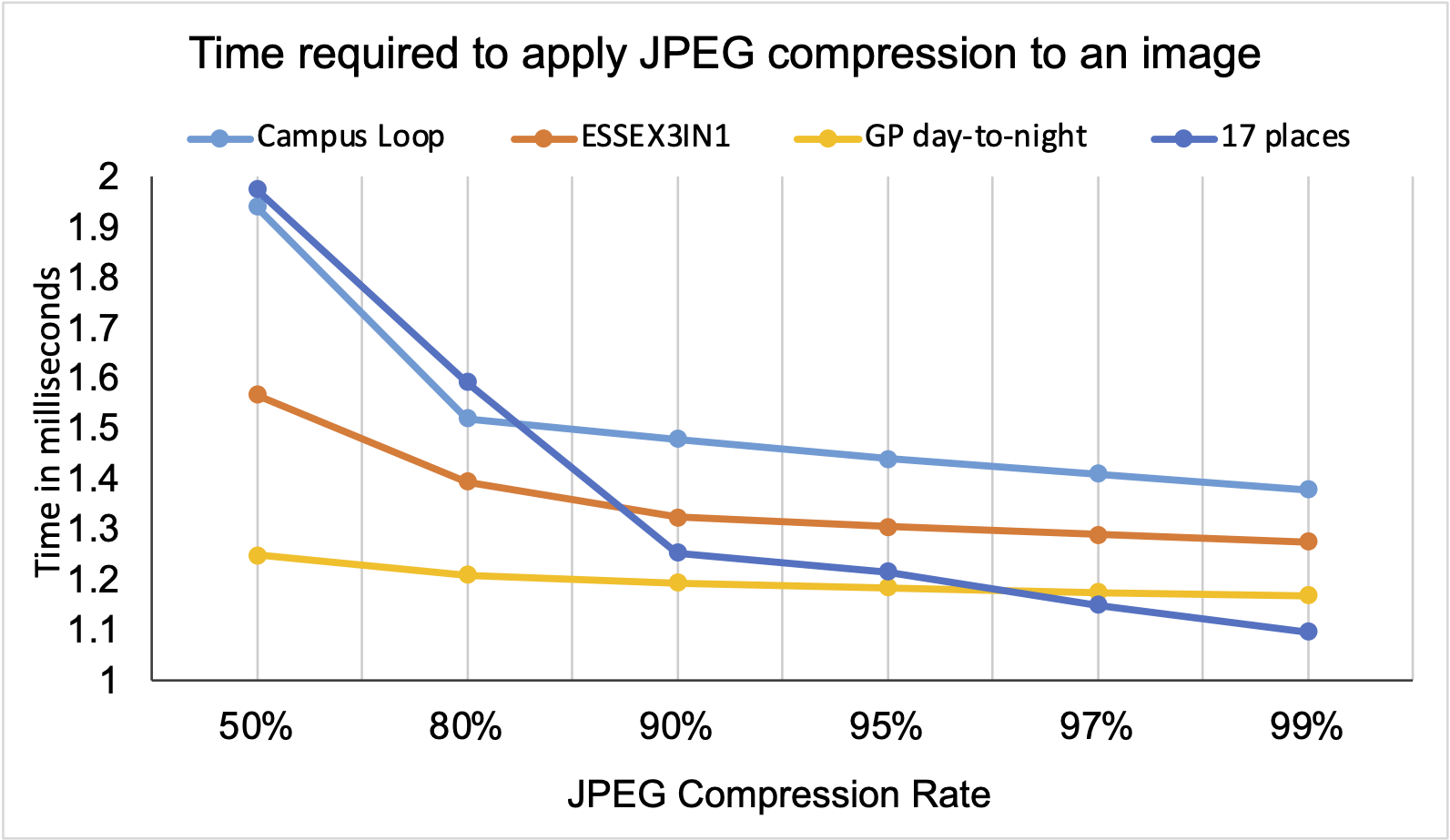} 
                
            \end{tabular}
            \caption{The average time $t_{c}$ required to JPEG compress an image.}
            \label{average_compression_time}
            \end{figure}

         Fig. \ref{average_compression_time} shows the time $t_{c}$ required to apply JPEG compression to a given image for each of the four datasets tested. The time $t'_{c}$ required to JPEG compress an entire sequence of images of length \textbf{K} can be computed using equation (\ref{eq:compression_time}):
        
        \begin{equation}\label{eq:compression_time}
            t'_{c} = t_c * K 
        \end{equation} 
        \begin{equation}\label{eq:total_time}
            t_{total} = t_{VPR} + t'_{c}
        \end{equation} 
    
        In Table \ref{table:time_comparison} a comparison is provided between the $t_{VPR}$ in a scenario where ESSEX3IN1 dataset is uniformly and non-uniformly JPEG compressed. Due to page limitations, we have only included the results for this dataset to show that the amount of time required to perform VPR can be drastically reduced if the sequence of query images is compressed to the same quality as the map. As JPEG compression is an extremely fast operation to perform, $t_{total}$ (refer to equation (\ref{eq:total_time})) is not considerably higher than $t_{VPR}$. The results presented in Table \ref{table:time_comparison} show that $t_{total}$ always benefits from using a shorter sequence length \textbf{K}, thus it is desirable to JPEG compress the query images to have the same quality as the map, as it facilitates more efficient place matching performance. However, we note some cases where the sequence length of a VPR technique is decreased on non-uniformly JPEG compressed data (refer to Fig. \ref{sequencelengthgraph} and Fig. \ref{non-uniform_compression}), more specifically HybridNet on Campus Loop (0\% query and 99\% map), RegionVLAD on Gardens Point day-to-night (0\% query, 99\% map) and HybridNet on 17 places (0\% query, 95\% map). In these scenarios, the sequence of query images should not be compressed to the same quality of the map, as it would increase the sequence length required to achieve maximum accuracy, while also increasing $t_{VPR}$.
        
        
            
            
    \section{Conclusions} \label{conclusion}
    To compensate for the reduction in VPR performance resulted from introducing high levels of JPEG compression, this paper introduces sequence-based filtering in several well-established single-frame-based VPR techniques. The sequence length that results in a perfect place matching performance is reported for every descriptor, throughout the entire spectrum of JPEG compression. To facilitate decentralised VPR applications where the limited bandwidth can impede the VPR process, the amount of data required to be transferred by each descriptor is analysed. Moreover, an investigation of the time required to perform VPR is provided. Our results show that a JPEG compressed image is often smaller in size when compared with an image descriptor and should be transmitted instead, in scenarios where limited bandwidth is available for VPR. Moreover, our experiments conclude that it is often advantageous compressing the query images to the same quality of the map, to perform more efficient VPR in changing environments.
    
    

    Future work can explore methods to optimise VPR techniques for JPEG imagery. These can include re-training/re-calibrating VPR techniques specifically for handling highly JPEG compressed data. Moreover, it would be interesting to investigate a sequence-based video compression codec such as H.265 \cite{sze2014high}, as an alternative to sequence-based VPR.


    {
    \small
    \bibliographystyle{ieeetr}
    \bibliography{root}
    }
    
\end{document}